\documentclass[times,twocolumn,final]{elsarticle}

\usepackage{framed,multirow}

\usepackage{amssymb}
\usepackage{latexsym}
\usepackage{epsfig}
\usepackage{graphicx}
\usepackage{amsmath}
\usepackage{amssymb}
\usepackage{gensymb}
\usepackage{math}
\usepackage{hyperref}
\usepackage[]{algorithm2e}
\usepackage[lofdepth,lotdepth]{subfig}
\graphicspath{{./figures/}}

\usepackage{url}
\usepackage{xcolor}
\definecolor{newcolor}{rgb}{.8,.349,.1}

\usepackage[final]{review} 

\begin{document}
\begin{frontmatter}

\title{Neural Network Based Reinforcement Learning for Audio-Visual Gaze Control in Human-Robot Interaction}
\author[1,2]{St\'{e}phane Lathuili\`{e}re}
\author[1,2]{Benoit Mass\'{e}}
\author[1,2]{Pablo Mesejo}
\author[1,2]{Radu Horaud}

\address[1]{Inria Grenoble Rh\^{o}ne-Alpes, Montbonnot-Saint-Martin, France}
\address[2]{Univ. Grenoble Alpes, Saint-Martin-d'H\`{e}res, France}
%
%
%

\begin{abstract}
This paper introduces a novel neural network-based reinforcement learning approach for robot gaze control. Our approach enables a robot to learn and to adapt its gaze control strategy for human-robot interaction neither with the use of external sensors nor with human supervision. The robot learns to focus its attention onto groups of people from its own audio-visual experiences, independently of the number of people, of their positions and of their physical appearances. In particular, we use a recurrent neural network architecture in combination with Q-learning to find an optimal action-selection policy; we pre-train the network using a simulated environment that mimics realistic scenarios that involve speaking/silent participants, thus avoiding the need of tedious sessions of a robot interacting with people. Our experimental evaluation suggests that the proposed method is robust against parameter estimation, i.e. the parameter values yielded by the method do not have a decisive impact on the performance. The best results are obtained when both audio and visual information is jointly used. Experiments with the Nao robot  indicate that our framework is a step forward towards the autonomous learning of socially acceptable gaze behavior.

\end{abstract}

%
%
\end{frontmatter}



\pagestyle{plain}

\section{Introduction}
\label{intro}

In recent years, there has been a growing interest in human-robot interaction (HRI), a research field dedicated to designing, evaluating and understanding robotic systems able to communicate with people \cite{Goodrich2007}. The robotic agent must perceive humans and perform actions that, in turn, will have an impact on the interaction. For instance, it is known that the robot's verbal and gaze behavior has a strong effect on the turn-taking conduct of the participants \cite{Skantze2014}. Traditionally, HRI has been focused on the interaction between a single person with a robot. However, robots are increasingly part of groups and teams, $e.g.$ performing delivery tasks in hospitals \cite{Ljungblad2012} or working closely alongside people on manufacturing floors \cite{Sauppe2015}. In the case of the gaze control problem in a multi-person scenario, the fact of focusing on only one person would lead to omit important information and, therefore, to make wrong decisions. Indeed, the robot needs to follow a strategy to maximize useful information, and such a strategy is difficult to design for two main reasons. First, handling all the possible situations with a set of handcrafted rules would be laborious and most-likely sub-optimal, especially when combining several sensors.  Second, the robot needs to be able to adapt its strategy to currently available data, as provided by its sensors, cameras and microphones in our case. For instance, if a companion robot enters a room with very bad acoustic conditions, the strategy needs to be adapted by decreasing the importance given to audio information.




\begin{figure*}[t!]
  \centering
  \includegraphics[width=0.95\textwidth]{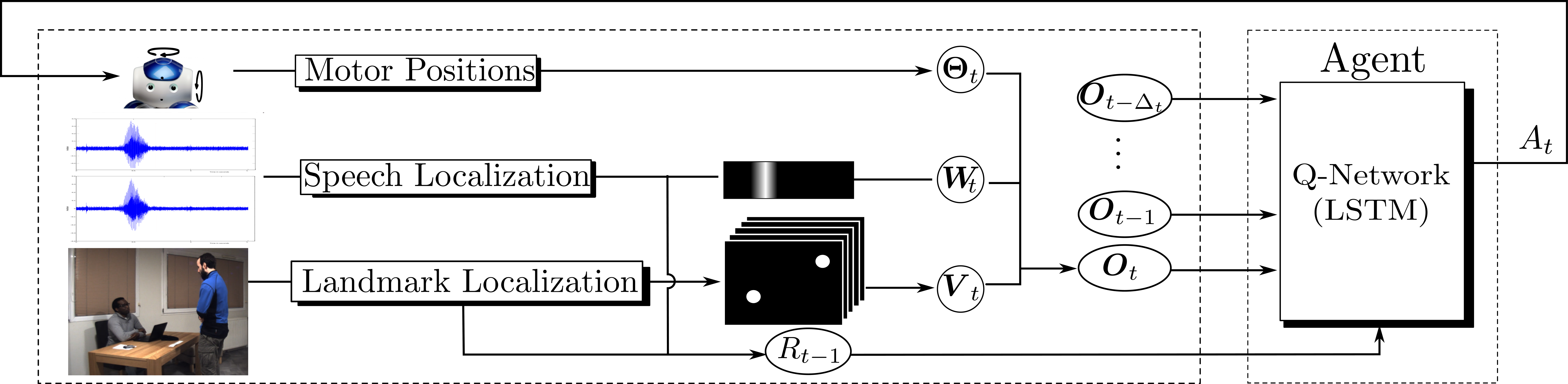}
    \label{fig:pipeline}
    \caption{Overview of the proposed deep RL method for controlling the gaze of a robot. At each time index $t$, audio and visual data are represented as binary maps which, together with motor positions, form the set of observations $\Ovect_t$. A motor action $A_t$ (rotate the head left, right, up, down, or stay still) is selected based on past and present observations via maximization of current and future rewards. The rewards $R$ are based on the number of visible persons as well as on the presence of speech sources in the camera field of view. We use a deep Q-network (DQN) model that can be learned both off-line and on-line. Please refer to Section~\ref{sec:model} and Section~\ref{QNetwork}  for the mathematical notations and detailed problem formulation.} 
\label{fig:pipeline}
\end{figure*}

In this paper, we consider the general problem of gaze control, with the specific goal of finding good policies to control the orientation of a robot head during informal group gatherings. In particular, we propose a methodology for a robotic system to be able to autonomously learn to focus its attention towards groups of people using audio-visual information. This is a very important topic of research since  perception requires not only making inferences from observations, but also making decisions about where to look next. More specifically, we want a robot to learn to find people in the environment, hence maximize the number of people present in its field of view, and favor people who speak. We believe this could be useful in many real scenarios, such as a conversation between a companion robot and a group of persons, where the robot needs to learn to look at people, in order to behave properly. The reason for using multiple sources of information can be found in recent HRI research suggesting that no single sensor can reliably serve robust interaction \cite{Pourmehr2017}. Importantly, when it comes to the employment of several sensing modalities in complex social interactions, it becomes difficult to implement an optimal policy based on handcrafted rules that take into consideration all possible situations that may occur. On the contrary, we propose to follow a data-driven approach to face such complexity. In particular, we propose to tackle this problem using a reinforcement learning (RL) approach \cite{Sutton1998}. RL is a machine learning paradigm in which agents learn by themselves by trial-and-error to achieve successful strategies.  As opposed to supervised learning, there is no need for optimal decisions at training time, only a way to evaluate how good a decision is: a reward. This paradigm, inspired from behavioral psychology, may enable a robot to autonomously learn a policy that maximizes accumulated rewards. In our case, the agent, a robot companion, autonomously moves its head depending on its knowledge about the environment. This knowledge is called the \textit{agent state}, and it is defined as a sequence of audio-visual observations, motor readings, actions, and rewards. In practice the optimal policy for making decisions is learned from the reward computed using detected faces of participants and sound sources being localized. The use of annotated data is not required to learn the best policy as the agent learns autonomously by trial-and-error in an unsupervised manner. Moreover, using our approach, it is not necessary to make any assumption about the number of people as well as their locations in the environment. 

The use of RL techniques presents several advantages. First, training using optimal decisions is not required since the model learns from the reward obtained from previous decisions. The reward may well be viewed as a feedback signal that indicates how well the robot is doing at a given time step. Second, the robot must continuously make judgments so as to select good actions over bad ones. In this sense, the model can keep training at test time and hence it benefits from a higher adaptation ability.  Finally, we avoid recourse to an annotated training set or calibration data. In our opinion, it seems entirely natural to use RL techniques to ``educate" a robot, since recent neuro-scientific studies have suggested that reinforcement affects the way infants interact with their environment, including what they look at \cite{Arcaro2017}, and that gazing at faces is not innate, but that environmental importance influences the gazing behavior.  


\addnote[contrib]{1}{The contributions of this paper are the followings.
First, robot gaze control is formulated as a reinforcement learning problem, allowing the robot to autonomously learn its own gaze control strategy from multimodal data.
Second, we use \textit{deep} reinforcement learning to model the action-value function, and suggest several architectures based on LSTM (a recurrent neural network model) that allow us to experiment with both early- and late-fusion of audio and visual data.  Third, we introduce a simulated environment that enables us to learn the proposed deep RL model without the need of spending hours of tedious interaction. Finally, by experimenting with both a publicly available dataset and with a real robot, we provide empirical evidence that our method achieves state-of-the-art performance. 
}

\section{Related Work}
\label{related}

Reinforcement Learning has been successfully  employed in different domains, including robotics \cite{Kober2013}. The RL goal is to find a function, called a policy, which specifies which action to take in each state, so as to maximize some function ($e.g.$, the mean or expected discounted sum) of the sequence of rewards. Therefore, learning the suitable policy is the main challenge, and there are two main categories of methods to address it. First, policy-based methods define a space from the set of policies, and sample policies from this space. The reward is then used, together with optimization techniques, \emph{e.g.} gradient-based methods, to increase the quality of subsequent sampled policies \cite{williams1992}. Second, value-based methods consist in estimating the expected reward for the set of possible actions, and the actual policy uses this value function to decide the suitable action, \emph{e.g.} choose the action that maximizes the value-function. In particular, popular value-based methods include Q-learning \cite{Watkins1992} and its deep learning extension, Deep Q-Networks (or DQNs) \cite{mnih-atari-2013}. 

There are several RL-based HRI methods relevant to our work. In \cite{ghadirzadeh2016sensorimotor} an RL algorithm is used for a robot to learn to play a game with a human partner. The algorithm uses vision and force/torque feedback to choose the motor commands. The uncertainty associated with human actions is modeled via a Gaussian process model, and Bayesian optimization selects an optimal action at each time step. In \cite{mitsunaga2006robot} RL is employed to adjust motion speed, timing, interaction distances, and gaze in the context of HRI. The reward is based on the amount of movement of the subject and the time spent gazing at the robot in one interaction. As external cameras are required, this cannot be easily applied in scenarios where the robot has to keep learning in a real environment. Moreover, the method is limited to the case of a single human participant. Another example of RL applied to HRI can be found in \cite{thomaz2006reinforcement}, where a human-provided reward is used to teach a robot. This idea of interactive RL is also exploited in \cite{cruz2016multi} in the context of a table-cleaning robot. Visual and speech recognition are used to get advice from a parent-like trainer to enable the robot to learn a good policy efficiently. An extrinsic reward is used in \cite{rothbucher2012robotic} to learn how to point a camera towards the active speaker in a conversation. Audio information is used to determine where to point the camera, while the reward is provided using visual information: the active speaker raises a blue card that can be easily identified by the robot. The use of a multimodal deep Q-network (DQN) to learn human-like interactions is proposed in both \cite{qureshi16gain} and \cite{QureshiNYI17}. The robot must choose an action to shake hands with a person. The reward is either negative, if the robot tries unsuccessfully to shake hands, positive, if the hand-shake is successful, or null otherwise. In practice, the reward is obtained from a sensor located in the hand of the robot and it takes fourteen training days to learn this skill successfully. To the best of our knowledge, the closest work to ours is \cite{vasquez16maintain} where an RL approach learns good policies to control the orientation of a mobile robot during social group conversations. The robot learns to turn its head towards the speaking person. However, their model is learned on simulated data that are restricted to a few predefined scenarios with static people and a predefined spatial organization of the groups. 

Gaze control has been addressed in the framework of sensor-based servoing. In \cite{Bennewitz2005} an ad-hoc algorithm is proposed to detect, track, and involve multiple persons into an interaction, combining audio-visual observations.
In a multi-person scenario, \cite{Ban17} investigated the complementary nature of tracking and visual servoing that enables the
system to track several persons and to visually control the gaze such as to keep
a selected person in the camera field of view.  Also, in \cite{yun2017gaze}, a system for gaze control of socially interactive robots in multiple-person scenarios is presented. This method requires external sensors to locate human participants. 

\addnote[contribccl]{1}{However, in opposition to all these works, we aim at learning an optimal gaze control behavior using minimal supervision provided by a reward function, instead of adopting a handcrafted gaze control strategy. Importantly, our model requires neither external sensors nor human intervention to compute the reward, allowing the robot to autonomously learn where to gaze.}


\section{Reinforcement Learning for Gaze Control}
\label{sec:model}
\addnote[pbIntro]{1}{
We consider a robot task whose goal is to look at a group of people. Hence, the robot must learn by itself a gazing strategy via trials and errors. The desired robot action is to rotate its head (endowed with a camera and four microphones), such as to maximize the number of persons lying in the camera field-of-view. Moreover, the robot should prefer to look at speaking people instead of silent ones. 
The terms \textit{agent} and \textit{robot} will be used indistinctly.}


Random variables and their realizations are denoted with uppercase and lowercase letters, respectively. Vectors and matrices are in bold italic. At each time index $t$, the agent gathers motor $\Thetavect_t$, visual $\Vvect_t$, and audio $\Wvect_t$ observations and performs an action $A_t\in\mathcal{A}$ from an action set according to a policy $\pi$, i.e. controlling the two head motors such that the robot gazes in a selected direction. Once an action is performed, the agent receives a reward $R_t$, as explained in detail below.

Without loss of generality we consider the companion robot Nao whose head has two rotational degrees of freedom:
motor readings correspond to pan and tilt angles, $\Thetavect_t=(\mathit{\Theta}^{1}_{t},\mathit{\Theta}^{2}_{t})$. The values of these angles are relative to a reference head orientation, e.g. aligned with the robot body. This reference orientation together with the motor limits define the robot-centered \textit{motor field-of-view} (M-FOV).

 \addnote[Vmaps]{1}{
We use the multiple person detector of \cite{cao2017realtime} to estimate two-dimensional visual landmarks, i.e. image coordinates, for each detected person, namely the nose, eyes, ears, neck, shoulders, elbows, wrists, hip, knees and ankles, or a total of $J=18$ possible landmarks for each person. Based on the detection of these landmarks, one can determine the number of (totally or partially) observed persons, $N_t$, as well as the number of observed faces, $F_t$. Notice that in general the number of faces that are present in the image (i.e. detection of nose, eyes or ears) may be smaller than the number of detected persons. Since the camera is mounted onto the robot head, the landmarks are described in a head-centered reference system. Moreover, these landmarks are represented by $J$ binary maps of size $K_v \times L_v$, namely $\Vvect_t\in \{0,1\}^{K_v\times L_v\times J}$, where 1 (resp. zero) corresponds to the presence (resp. absence) of a landmark. Notice that this representation gathers all the detected landmarks associated with the $N_t$ detected persons.} 

 \addnote[Amaps]{1}{
Audio observations are provided by the multi audio-source localization method described in \cite{li2017multiple}. 
Audio observations are also represented with a binary map of size $K_a \times L_a$, namely $\Wvect_t\in \{0,1\}^{K_a\times L_a}$. A map cell is set to 1 if a speech source is detected at that cell and 0 otherwise. The audio map is robot-centered and hence it remains fixed whenever the robot turns its head. Moreover, the audio map spans an \textit{acoustic field-of-view} (A-FOV), which is much wider than the \textit{visual field-of-view} (V-FOV), associated with the camera mounted onto the head. The motor readings allow to estimate the relative alignment between the audio and visual maps and to determine whether a speech source lies within the visual field-of-view or not. This is represented by the binary variable $\Sigma_t\in\{0,1\}$, such that $\Sigma_t =1$ if a speech source lies in the visual field-of-view and $\Sigma_t =0$ if none of the speech sources lies inside the visual field-of-view.
}

Let $\Ovect_t=\{\Thetavect_t,\Vvect_t,\Wvect_t\}$ and let $\Svect_t=\{\Ovect_1, \dots, \Ovect_t\}$ denote the state variable.
Let the set of actions be defined by $\mathcal{A}=\{ \varnothing~, \leftarrow~, \uparrow~, \rightarrow~, \downarrow\}$, namely either remain in the same position or turn the head by a fixed angle in one of the four cardinal directions. \addnote[rewards]{1}{We propose to define the reward $R_t$ as follows:
\begin{align}
  R_t=F_{t+1}+\alpha\Sigma_{t+1}, \label{rewarddef}
\end{align}
where $\alpha\geq 0$ is an adjustment parameter. 
Large $\alpha$ values return high rewards when speech sources lie within the camera field-of-view. We consider two types of rewards which are referred to in Section~\ref{exper} as \emph{Face$\_$reward} ($\alpha=0$) and  \emph{Speaker$\_$reward} ($\alpha =1$). Notice that the number of observed faces, $F_{t}$, is independent of the speaking state of each person.  Upon the application at hand, the value of $\alpha$ allows one to weight the importance given to speaking persons.}

In RL, the model parameters are learned on sequences of states, actions and rewards, called episodes.
At each time index $t$, an optimal action $A_t$ should be chosen by maximizing the immediate and future rewards, $R_t,R_{t+1}, \dots, R_T$. We make the standard assumption that future rewards are discounted by a factor $\gamma$ that defines the importance of short-term rewards as opposed to long-term ones. We define the discounted future return $\bar{R}_t$ as the discounted sum of future rewards, $\bar{R}_{t}=\sum_{\tau=0}^{T-t}\gamma^\tau R_{\tau+t}$. If $\gamma=0$, $\bar{R}_t= R_t$ and, consequently, we aim at maximizing only the immediate reward whereas when $\gamma\approx1$, we favor policies that lead to better rewards in the long term. Considering a fixed value of $\gamma$, we now aim at maximizing $\bar{R}_{t}$ at each time index $t$. In other words, the goal is to learn a policy, $\pi(a_t,\svect_t)=P(A_t=a_t|\Svect_t=\svect_t)$ with $(a_t,\svect_t) \in \mathcal{A}\times \mathcal{S}$, such that if the agent chooses its actions according to the policy $\pi$, the expected $\bar{R}_{t}$ should be maximized. The Q-function (or the action-value function) is defined as the expected future return from state $\Svect_t$, taking action $A_t$ and then following any given policy $\pi$:
\begin{align}
  Q_\pi(\svect_t , a_t)=\mathbb{E}_\pi[\bar{R}_{t}|\Svect_t=\svect_t,A_t=a_t].
  \label{Eq1}
  \end{align}
Learning the best policy corresponds to the following optimization problem $Q^*(\svect_t,a_t)=\underset{\pi}{\max}[Q_\pi(\Svect_t=\svect_t, A_t=a_t)]$.
 The optimal Q-function obeys the identity known as the Bellman equation:
\begin{align}
  Q^*(\svect_t , a_t) & =\mathbb{E}_{\Svect_{t+1},R_t}\Big[R_t \nonumber \\
  &+\gamma~\underset{a}{\max}(Q^*(\Svect_{t+1}, a))\Big|\Svect_t=\svect_t,A_t=a_t\Big]
  \label{EqBellman}
  \end{align}
 This equation corresponds to the following intuition: if we have an estimator $Q^*(\svect_{t} , a_{t})$ for $\bar{R}_{t}$, the optimal action $a_t$ is the one that leads to the largest expected $\bar{R}_t$. The recursive application of this policy leads to equation \eqref{EqBellman}. A straightforward approach would consist in updating $Q$ at each training step $i$ with:
\begin{align}
  Q^{i}(\svect_t , a_t) &= \mathbb{E}_{\Svect_{t+1},R_t}\Big[R_t \nonumber \\
  &+\gamma~\underset{a}{\max}(Q^{i-1}(\Svect_{t+1}, a))\big|\Svect_t=\svect_t,A_t=a_t\Big]
  \label{EqBellmanTr}
  \end{align}
Following equation \eqref{EqBellmanTr}, we estimate each action-value $Q^i(\svect_t , a_t)$ given that we follow, for the next time steps, the policy implied by $Q^{i-1}$. In practice, we approximate the true Q function by a function whose parameters must be learned. In our case, we employ a network  $Q(\svect , a,\omegavect)$ parametrized by weights $\omegavect$ to estimate the Q-function  $Q(\svect , a,\omegavect)\approx Q^*(\svect , a)$. We minimize the following  loss:
\begin{align}
  \mathcal{L}(\omegavect_{i})=\mathbb{E}_{\Svect_t,\Avect_t,\Rvect_t,\Svect_{t+1}}\Big[(Y_{i-1}-Q(\Svect_t , \Avect_t,\omegavect_{i}))^2\Big]
  \label{lossDQN}
\end{align}
with $Y_{i-1}=R_t+\gamma~\underset{a}{\max}(Q(\Svect_{t+1}, a,\omegavect_{i-1}))$. This may be seen as minimizing the mean squared distance between approximations of the right- and left-hand sides of \eqref{EqBellmanTr}. 
In order to compute \eqref{lossDQN}, we sample quadruplets $(\Svect_t,\Avect_t,\Rvect_t,\Svect_{t+1})$ following the policy implied by $Q^{i-1}$:
\begin{align}
  a_t=\underset{a \in \mathcal{A}}{\argmax}Q(\svect_t , a,\omega_{i-1})
  \label{Eqaction}
\end{align}

However, instead of sampling only according to \eqref{Eqaction}, random actions $a_t$ are taken in $\epsilon$ percents of the time steps in order to explore new strategies. This approach is known as epsilon-greedy policy. $\mathcal{L}$ is minimized over $\omegavect_{i}$ by stochastic gradient descent. Refer to \cite{mnih2015human} for more technical details about the training algorithm.

\subsection{Neural Network Architectures for Q-Learning}
\label{QNetwork}
The Q-function is modeled by a neural network that takes as input
part of the state variable $\Svect_t$, that we define as $\Svect^{\Delta t}_t=\{\Ovect_{t-\Delta  t}...\Ovect_t\}$. The output is a vector of size $\# \mathcal{A}$ that
corresponds to each $Q_\pi(\svect^{\Delta t}_t , a_t), a_t \in \mathcal{A}$, where $Q_\pi(\svect^{\Delta t}_t , a_t)$ is built analogously to \eqref{Eq1}. Following \cite{mnih2015human}, the output layer is a fully connected layer (FCL) with linear activations. We propose to use the long short-term memory (LSTM)~\cite{lstmSchmidhuber} recurrent neural network to model the Q-function since recurrent neural networks are able to exhibit dynamic behavior for temporal sequences. LSTM are designed such as to prevent the back propagated errors from vanishing or exploding during training. We argue that LSTM is well suited for our task as it is capable of learning temporal dependencies better than other recurrent neural networks or than Markov models. In fact, our model needs to memorize the position and the motion of the people when it turns its head. When a person is not detected anymore, the network should be able to use previous detections back in time in order to predict the direction towards it should move. Batch normalization is applied to the output of LSTM. The $J$ channels of $\Vvect_t$ are flattened before the LSTM layers.

\begin{figure}[t!h!]
  \centering
  \begin{tabular}{cc}
  \subfloat[\emph{EFNet}]{
    \includegraphics[height=0.20\textwidth]{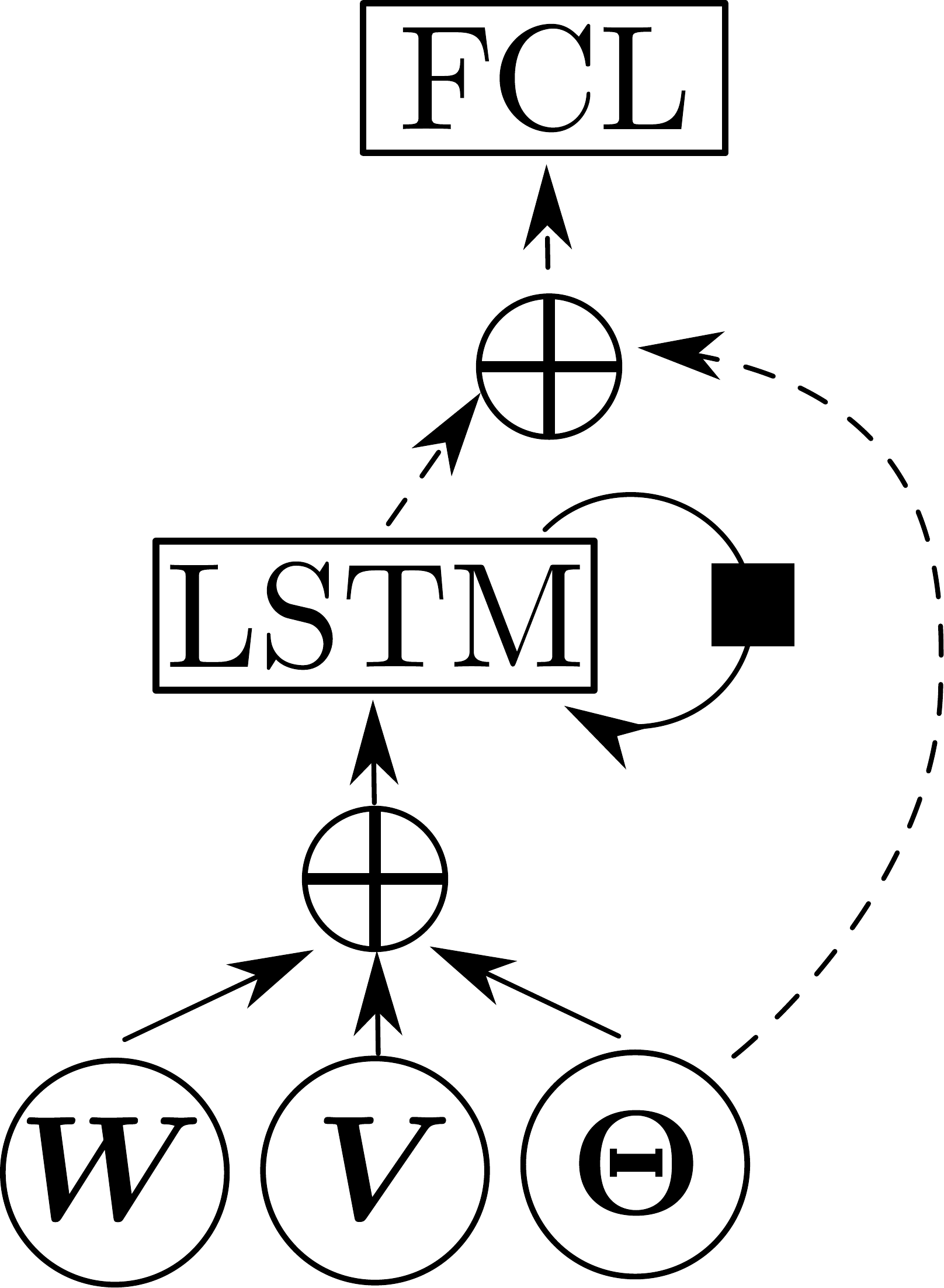}
 \label{fig:archsEarly}
  }&
  \subfloat[\emph{LFNet}]{
    \includegraphics[height=0.20\textwidth]{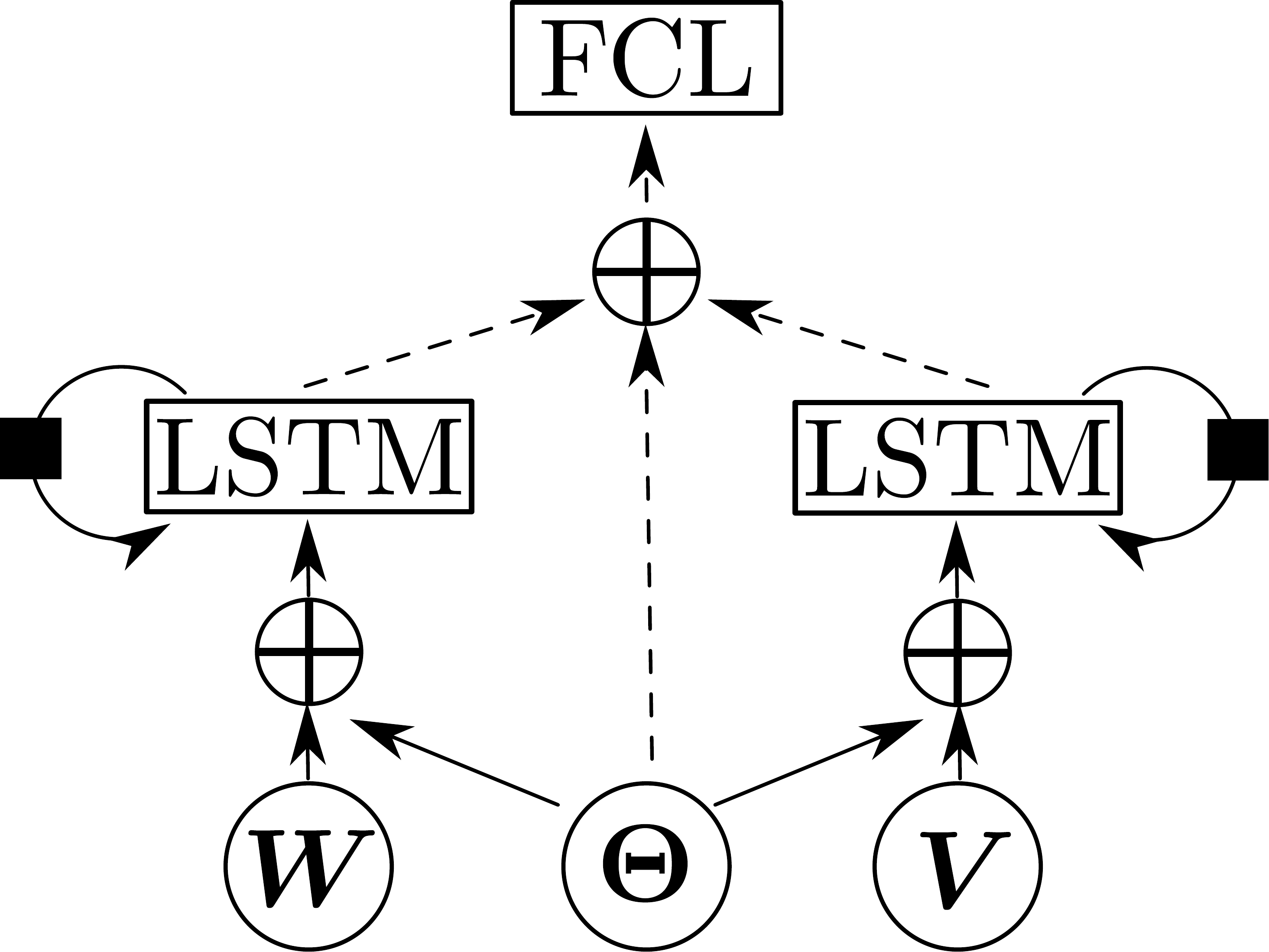}
 \label{fig:archsLate}
  }\\
  \subfloat[\emph{AudNet}]{
    \includegraphics[height=0.20\textwidth]{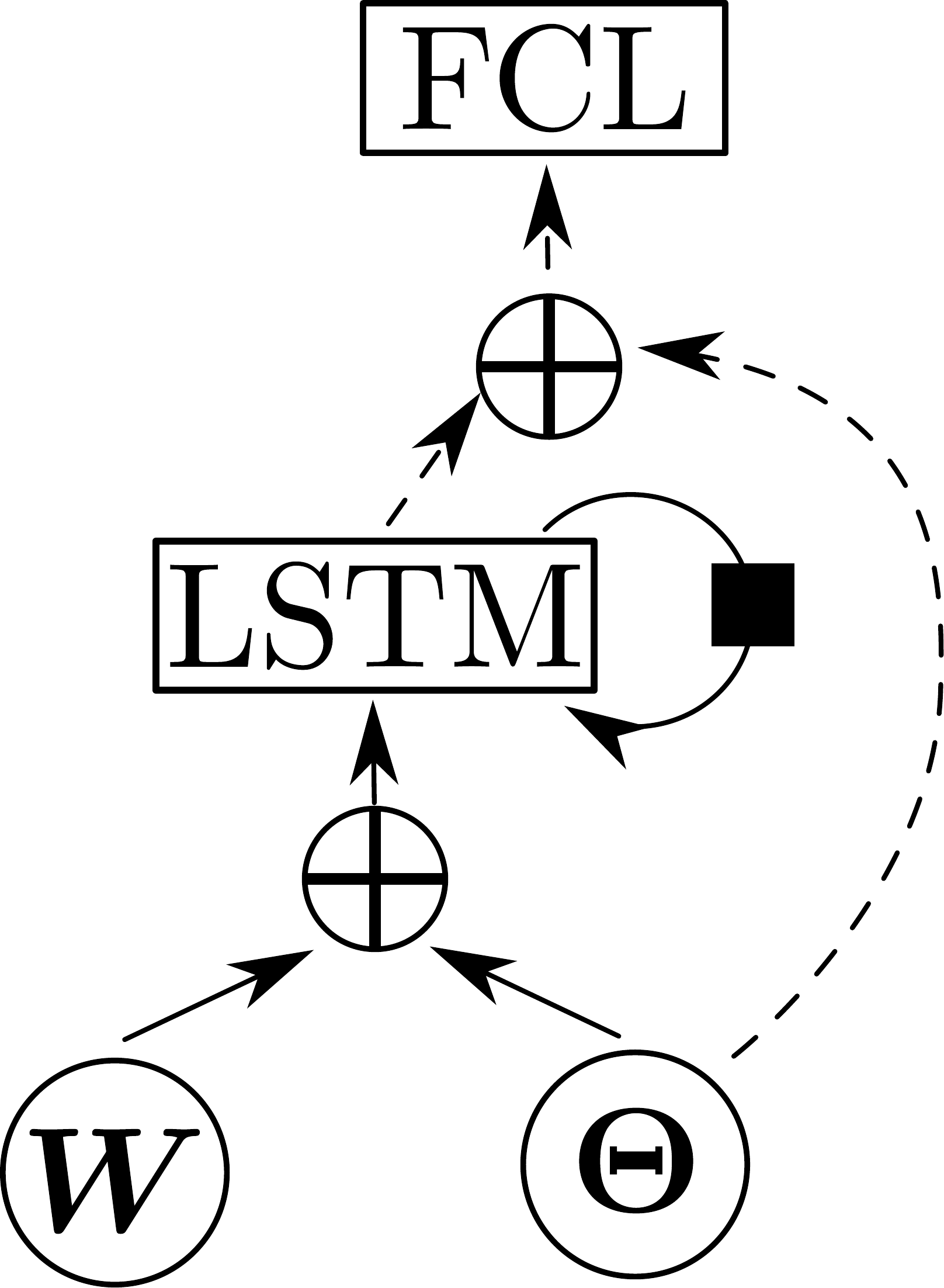}
 \label{fig:archsAudNet}
  }&
  \subfloat[\emph{VisNet}]{
    \includegraphics[height=0.20\textwidth]{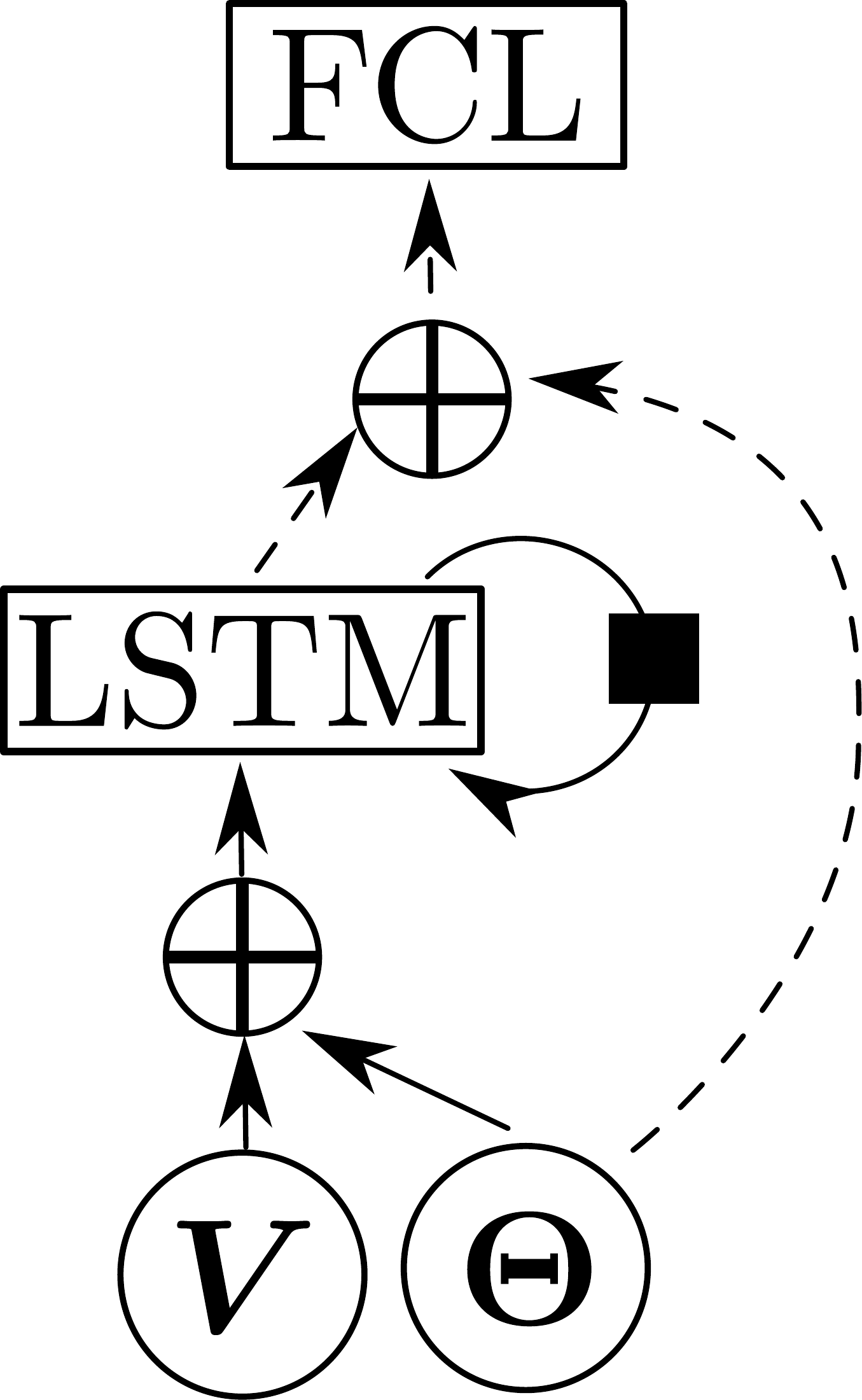}
\label{fig:archsVisNet}
  }
  \end{tabular}
  \caption{Proposed architectures to model the Q-function. Dashed lines indicate connections only used in the last time step. Black squares represent a delay of a single time step. Encircled crosses depict the concatenation of inputs. }
\label{fig:archs}
\end{figure}

\begin{figure}[h!]
  \centering
\includegraphics[width=0.45\textwidth]{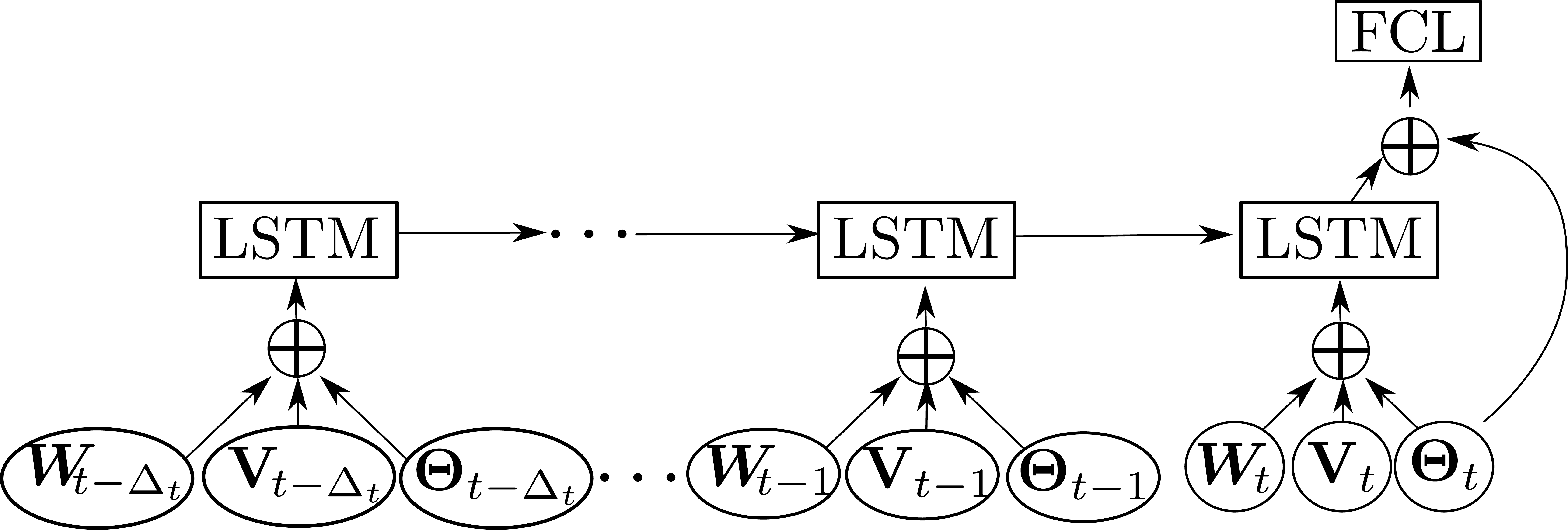}
 \caption{Unfolded representation of \emph{EFNet}  to better capture the sequential nature of the recurrent model. Encircled crosses depict the concatenation of inputs. }
\label{fig:unfoldedArchsEarly}
\end{figure}

Four different network architectures are described in this section and are evaluated in Section \ref{exper}. In order to evaluate when the two streams of information (audio and video) need to be fused, we propose to compare two architectures: early fusion (\emph{EFNet}) and late fusion (\emph{LFNet}). In early fusion, the audio and visual features  are  combined into a single representation before modeling time dependencies, e.g. Figure \ref{fig:archsEarly}. Conversely, in late fusion, audio and visual features are modeled separately before fusing them, e.g. Figure \ref{fig:archsLate}. In order to measure the impact of each modality, we also propose two more network architectures using either audio (\emph{AudNet}) or vision (\emph{VisNet}) information. Figure \ref{fig:archsAudNet} displays \emph{AudNet} and Figure \ref{fig:archsVisNet} displays  \emph{VisNet}. Figure \ref{fig:archs} employs a compact network representation where time is not explicitly shown, while Figure \ref{fig:unfoldedArchsEarly} depicts the unfolded representation of \emph{EFNet} where each node is associated with one particular time instance. Both figures follow the graphical representation used in \cite{goodfellow2016deep}.

\begin{figure}[h!]
  \centering
\includegraphics[width=0.50\columnwidth]{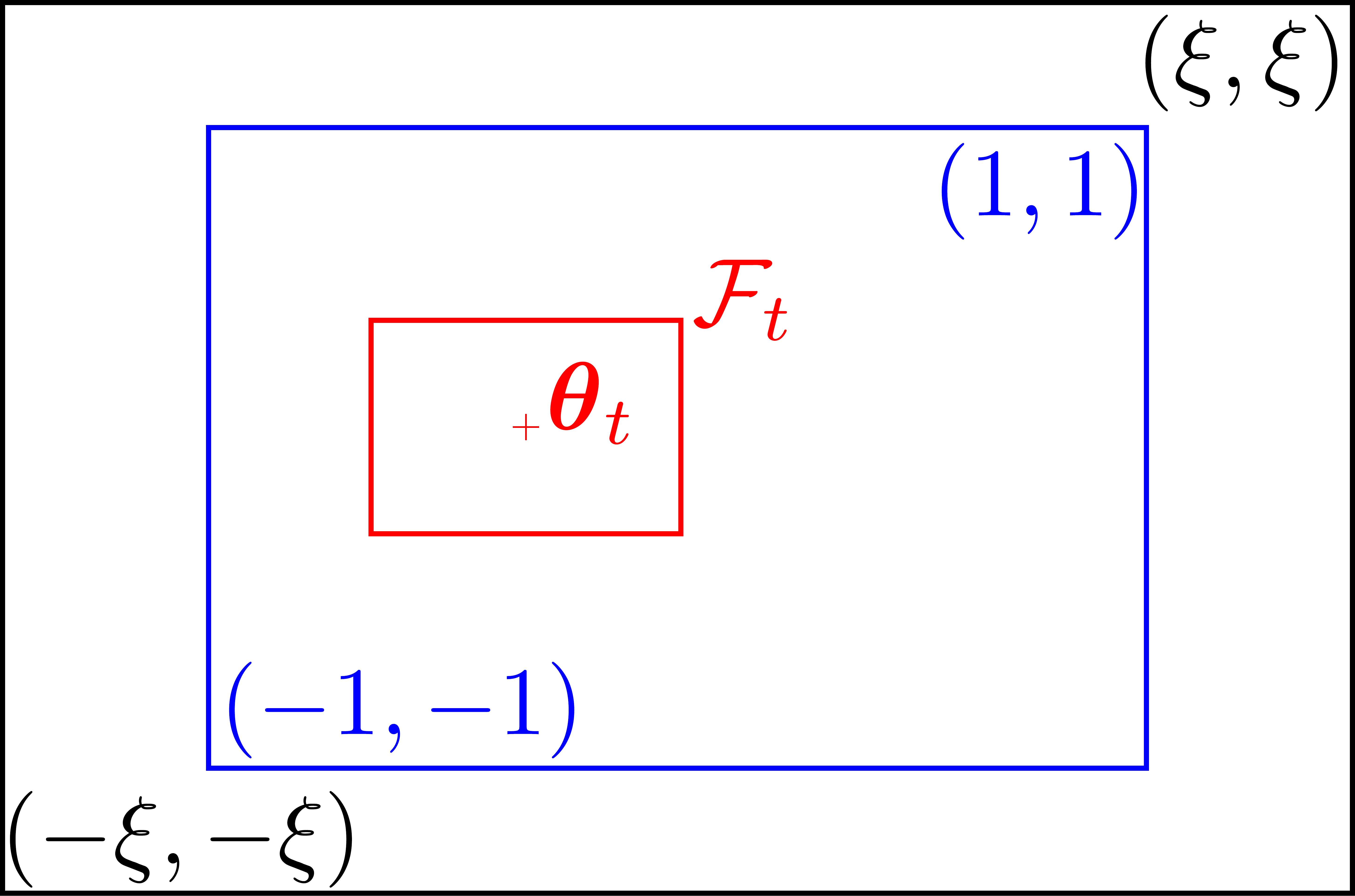}
 \caption{Diagram showing all fields used in the proposed simulated environment. The robot's field of view (in red) can move within the reachable field (in blue), whereas the participants can freely move within a larger field (in black).}
\label{fig:fieldsSimulated}
\end{figure}

\subsection{Pretraining on Simulated Environment}
\label{subsec:simulated}
\label{sec:synEnv}
\begin{figure}[t!]
  \centering
  \begin{tabular}{cc}
  \includegraphics[width=0.46\columnwidth]{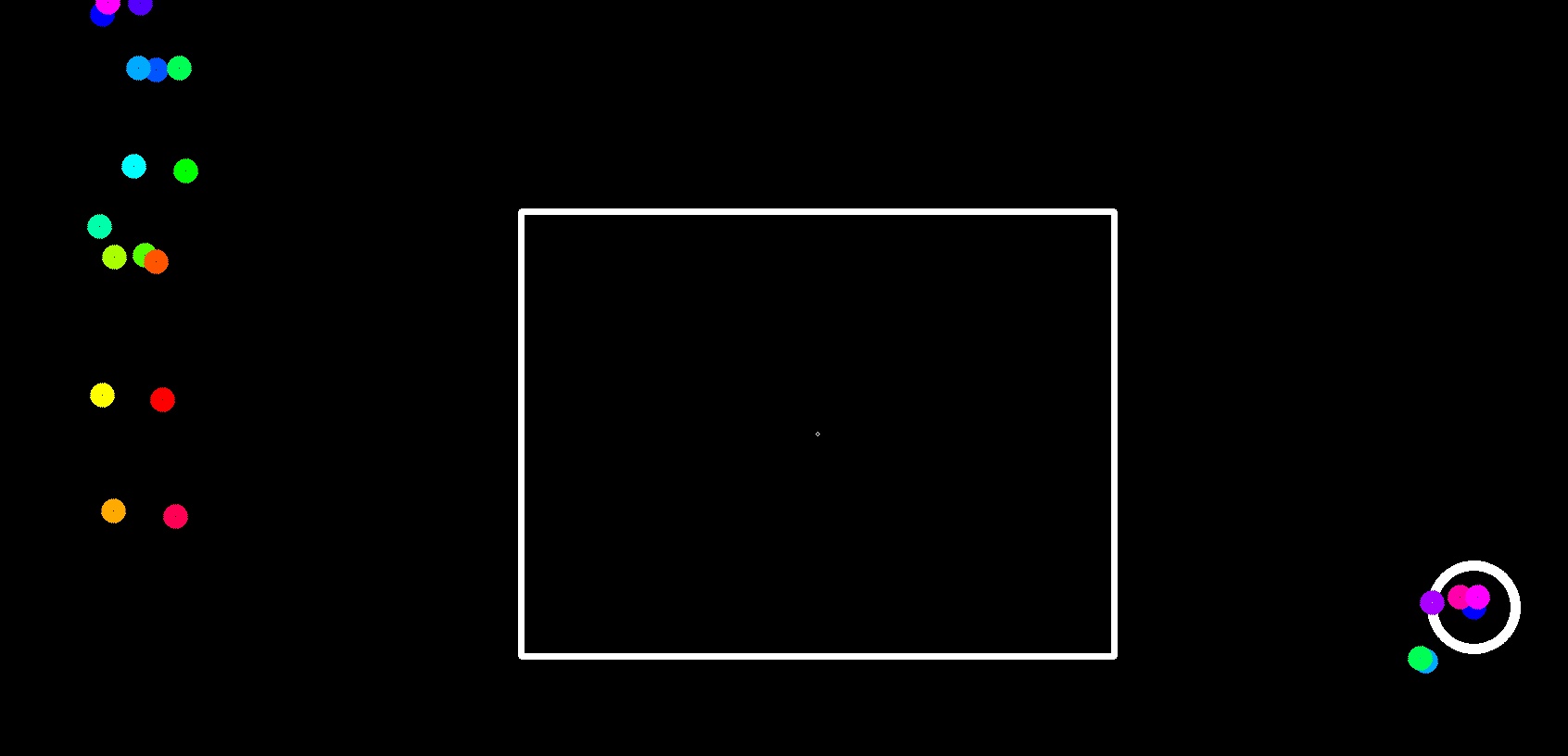}&
  \includegraphics[width=0.46\columnwidth]{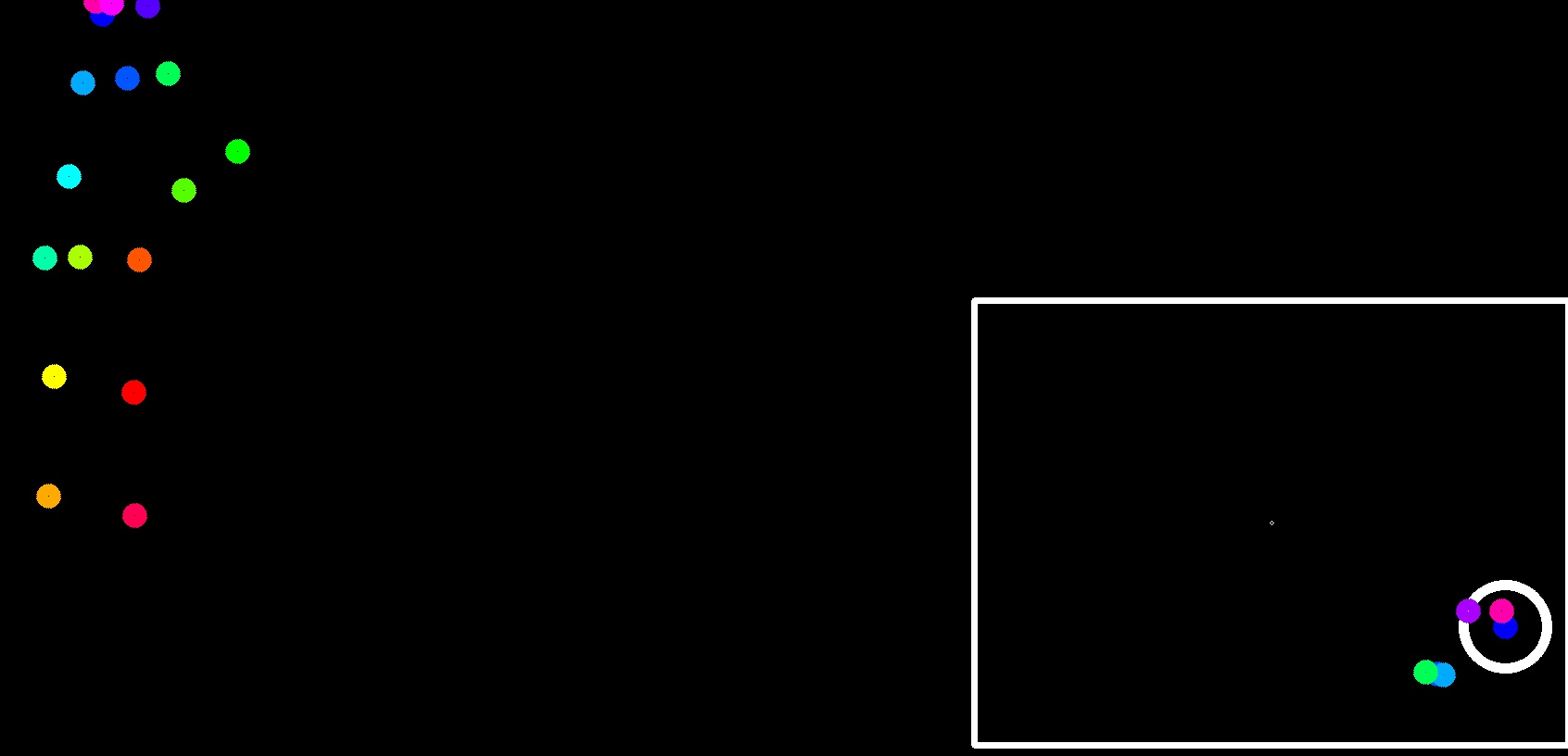}\\
  \includegraphics[width=0.46\columnwidth]{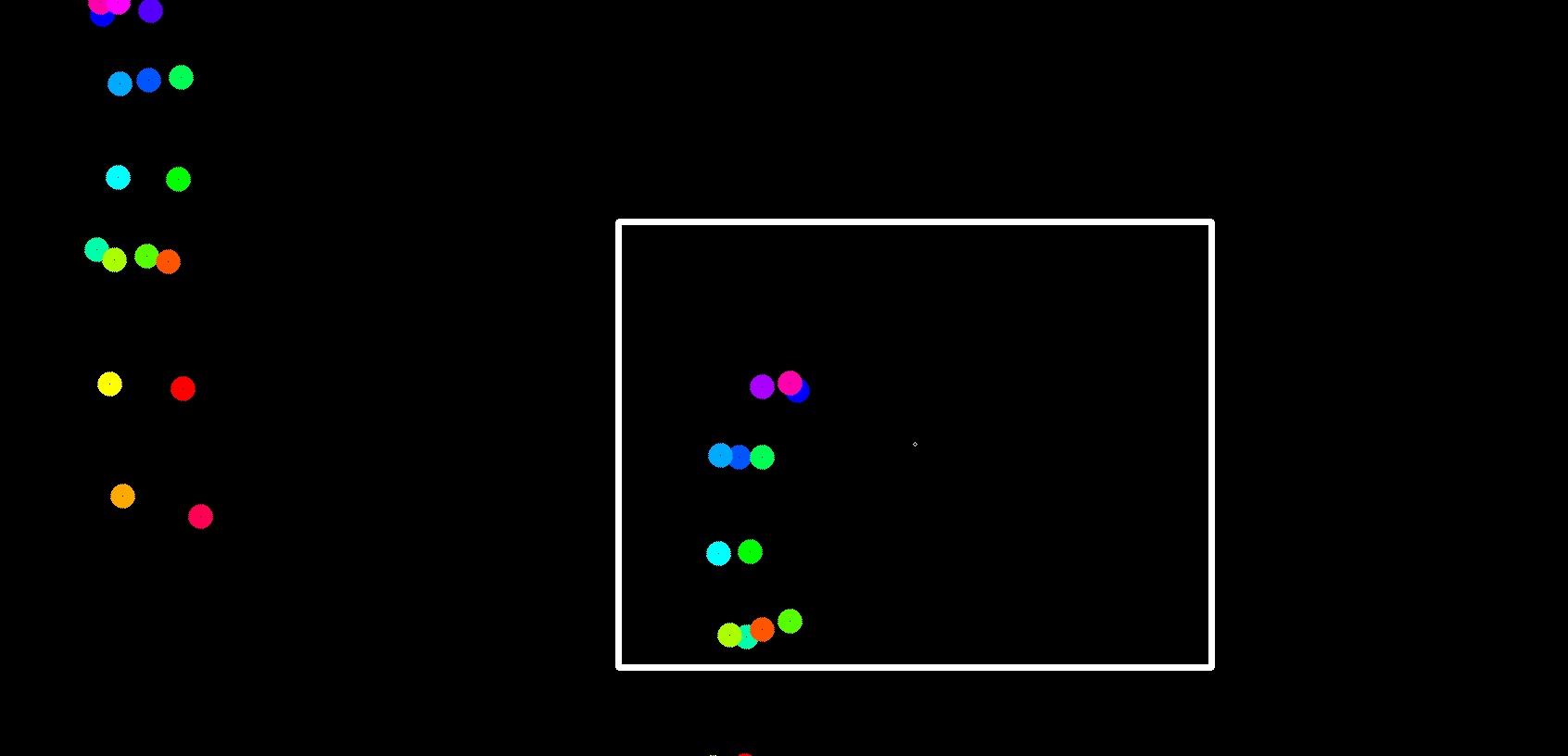}&
  \includegraphics[width=0.46\columnwidth]{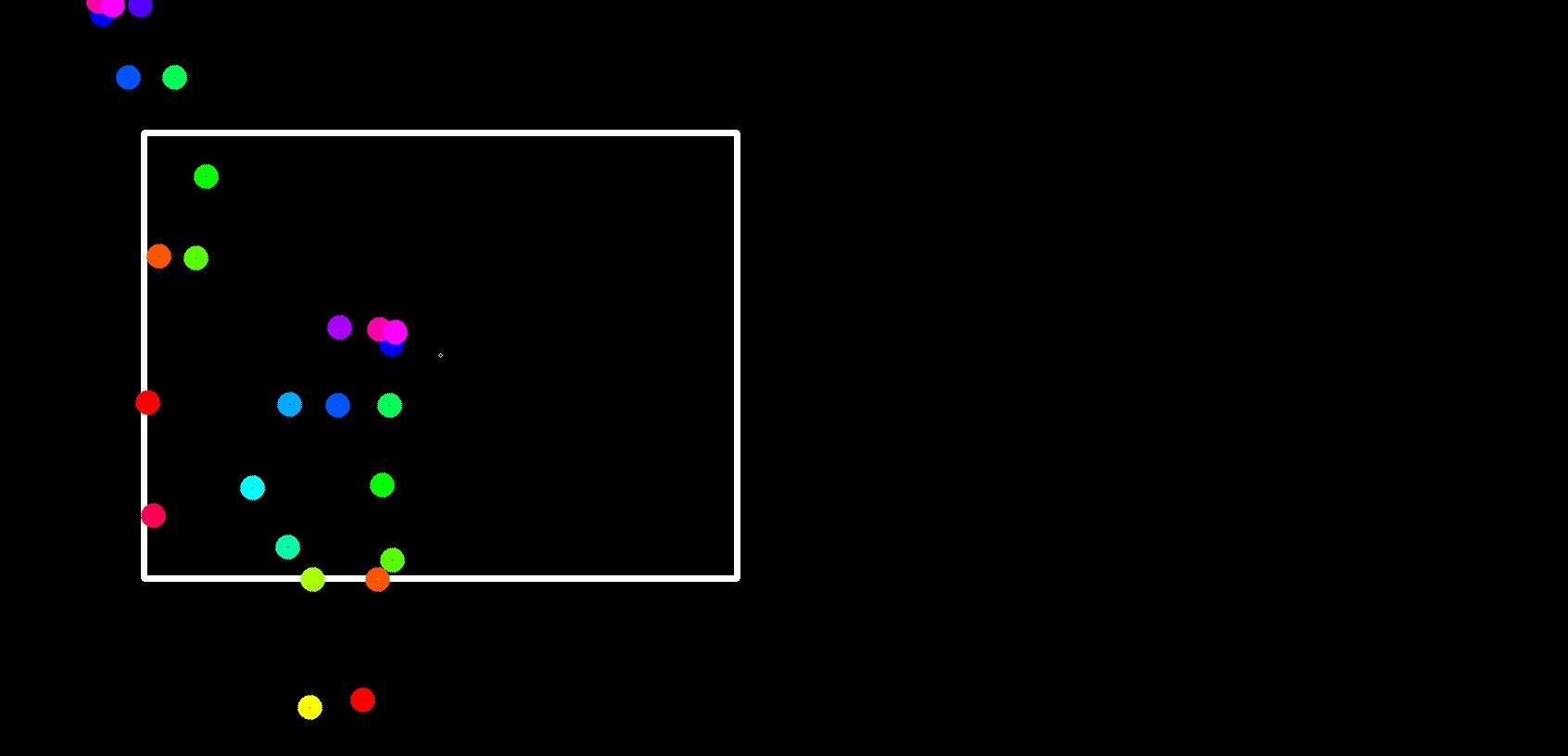}\\
  \includegraphics[width=0.46\columnwidth]{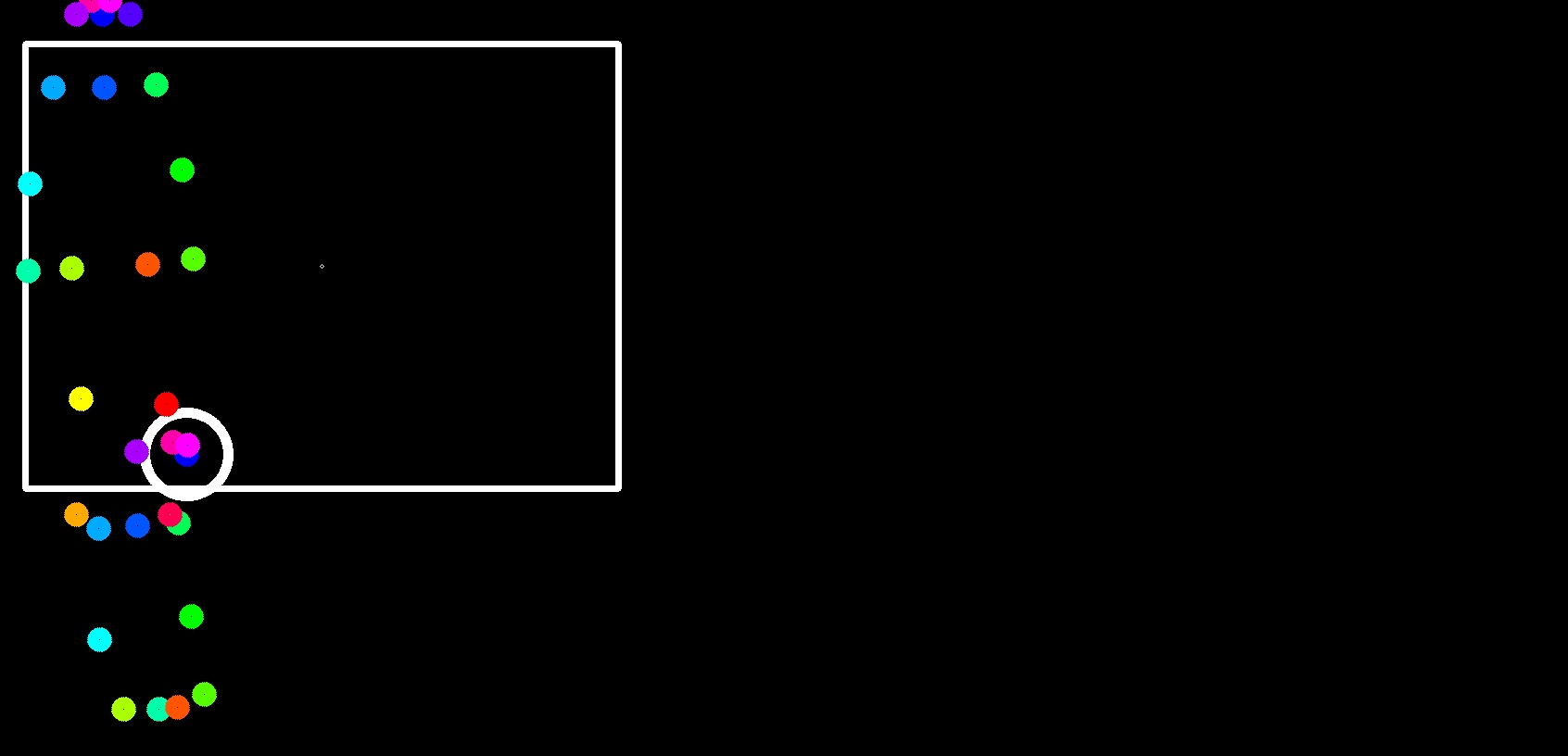}&
  \includegraphics[width=0.46\columnwidth]{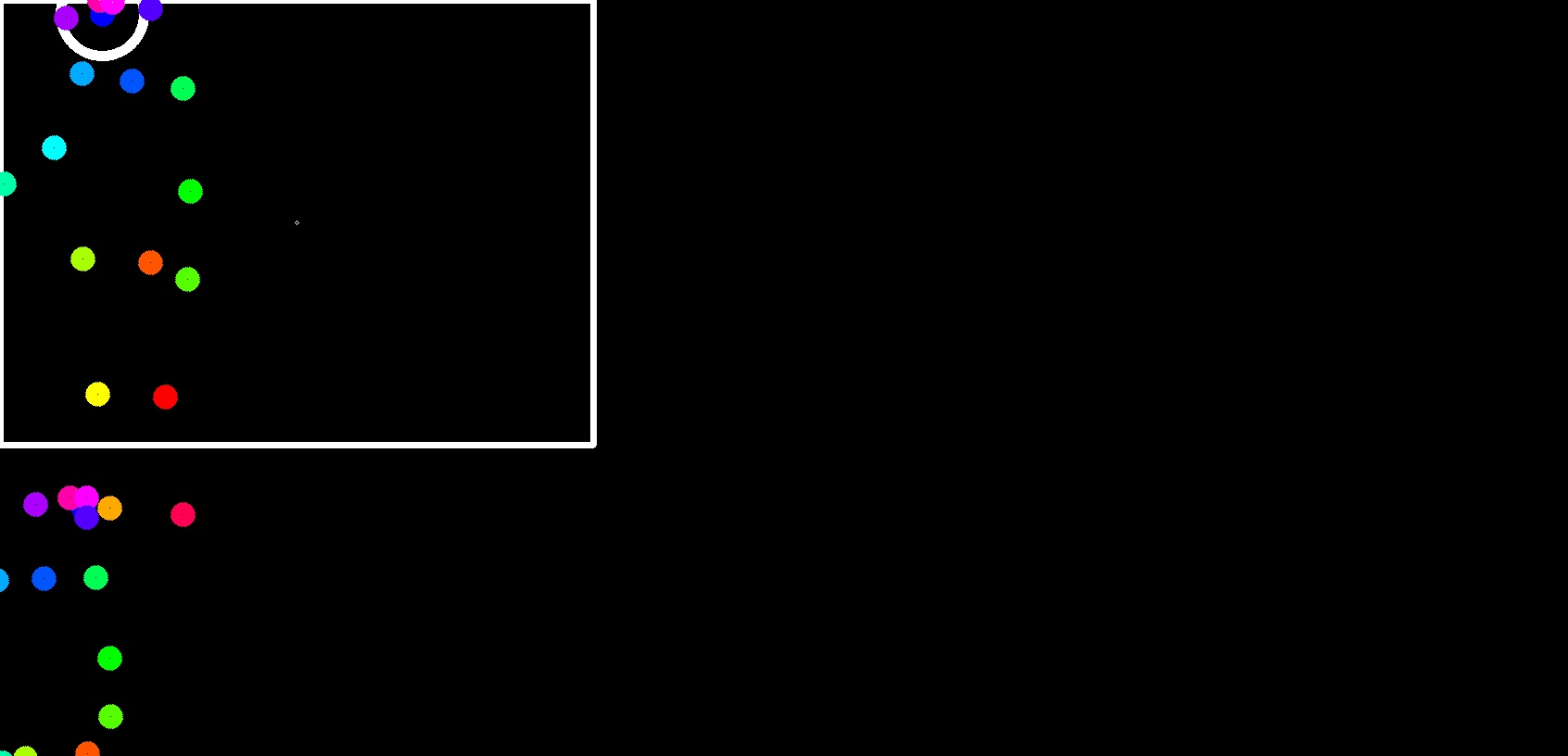}\\
  \includegraphics[width=0.46\columnwidth]{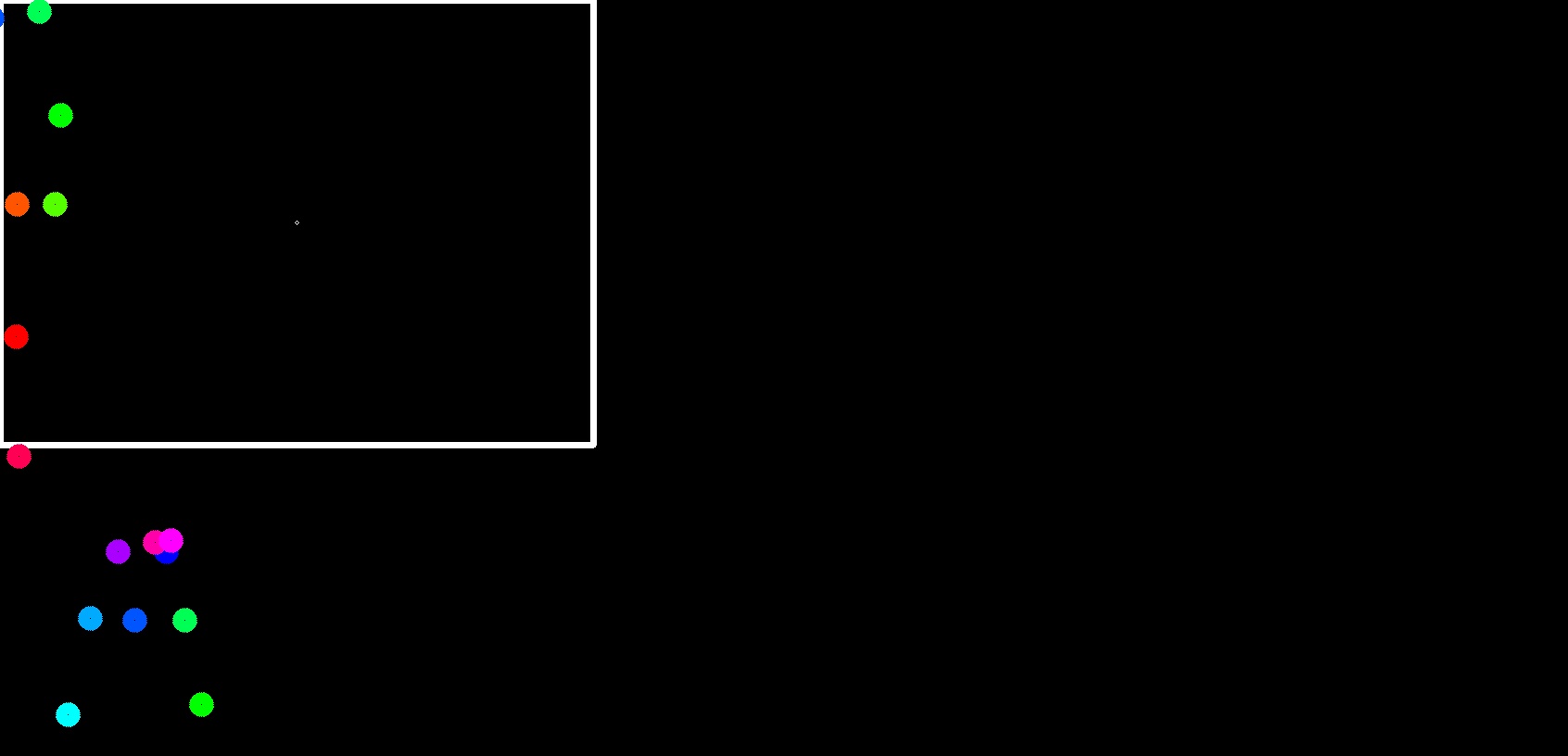}&
  \includegraphics[width=0.46\columnwidth]{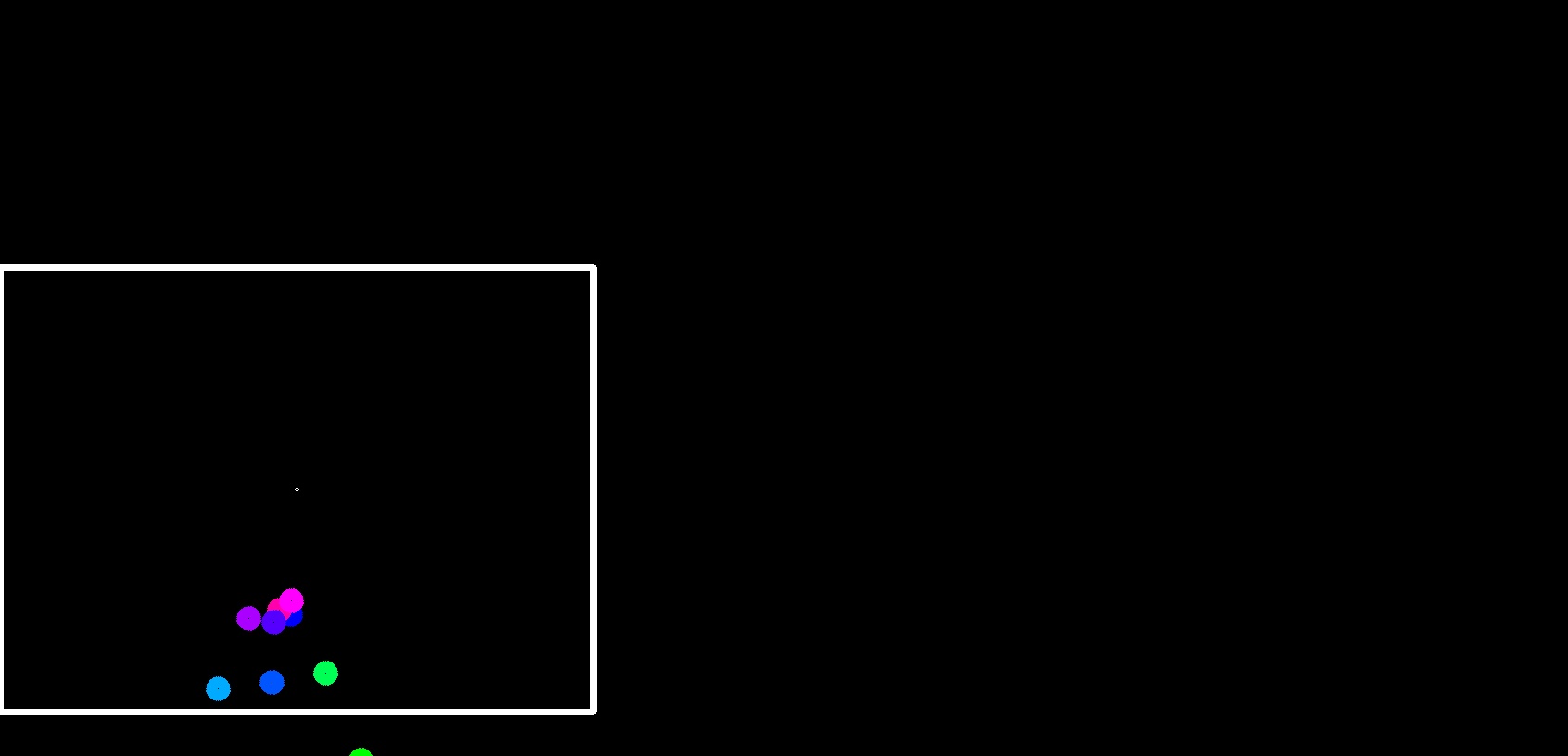}
  \end{tabular}
  \caption{Illustrative sequence taken from the simulated environment and employed to pretrain our neural network-based RL approach. The moving square represents the \textit{camera field-of-view} $\mathcal{F}_t$ of the robot. The colored circles represent the joints of a person in the environment. The large white circle represents a person speaking and, therefore, producing speech that can be detected by the speech localization system. Frames are displayed from top to bottom and left to right.}
\label{fig:simulatedEnvironment}
\end{figure}

Training from scratch a DQN model can take a long time (in our case $\sim$150000 time steps to converge), and training directly on a robot would not be convenient for two reasons. First, it would entail a long period of training, since each physical action by the robot takes an amount of time that cannot be reduced neither by code optimization nor by increasing our computational capabilities. Second, in the case of HRI, participants would need to move in front of the robot for several hours or days (like in \cite{qureshi16gain}). For these two reasons, we propose to use a transfer learning approach. The Q-function is first learned on a simulated environment, where we simulate people moving and talking, and it is then used to initialize the network employed by the robot. Importantly, the network learned from this simulated environment can be successfully used in the robot without the need of fine-tuning in real data. In this simulated environment, we do not need to generate  images and  sound signals, but only the observations and rewards the Q-Network receives as input.

We consider that the robot can cover the field $[-1,1]^2$ by moving its head, but can only visually observe  the people within a small rectangular region $\mathcal{F}_t\subset [-1,1]^2$ centered in position vector $\Thetavect_t$. The audio observations cover the whole reachable region $[-1,1]^2$.  On each episode, we simulate one or two persons moving with random speeds and accelerations within a field $[-\xi,\xi]^2$ where $\xi>1$.  In other words, people can go to regions that are unreachable for the robot. For each simulated person in the current episode, we consider the position and velocity of their head at time $t$, $\hvect_t=(u^{h}_t,v^{h}_t)\in [-\xi,\xi]^2$ and $\dot{h}=(\dot{u}^{h}_t,\dot{v}^{h}_t)\in \mathbb{R}^2$, respectively.
At each frame, the person can keep moving, stay without moving, or choose another random direction. The details of the simulated environment generator are given in Algorithm \ref{alg:synth}.
In a real scenario, people can leave the scene so, in order to simulate this phenomenon, we consider two equally probable cases when a person is going out horizontally of the field ($v_t^{h}\notin[-\xi,\xi]$).
In the first case, the person is deleted and instantly recreated on the other side of the field ($v_{t+1}^{h}=-v^{h}_t$) keeping the same velocity ($\dot{v}^{h}_{t+1}=\dot{v}^{h}_t$). In the second case, the person is going back towards the center ($v^{h}_{t+1}=v^{h}_t$ and ($\dot{v}^{h}_{t+1}=-\dot{v}^{h}_t$)). A similar approach is used when a person is going out vertically except that we do not create new persons on top of the field because that would imply the unrealistic sudden appearance of new legs within the field. Figure \ref{fig:fieldsSimulated} displays a visual representation of the different fields (or areas) defined in our simulated environment, and  Figure \ref{fig:simulatedEnvironment} shows an example of a sequence of frames taken from the simulated environment and used during training. 

Moreover, in order to favor tracking abilities, we bias the person motion probabilities such that a person that is faraway from the robot head orientation has a low probability to move, and a person within \textit{camera field-of-view} has a high probability to move. Thus, when there is nobody in the \textit{camera field-of-view}, the robot cannot simply wait for a person to come in. On the contrary, the robot needs to track the persons that are visible. More precisely, we consider 4 different cases. First, when a person has never been seen by the robot, the person does not move. Second, when a person is in the robot field of view ($\hvect_t\in\mathcal{F}_t$), they move with a probability of $95\%$. Third, when the person is further than a threshold $\tau\in \mathbb{R}$ from the \textit{camera field-of-view} ($||\hvect_t - \Thetavect_t||_2>\tau$), the probability of moving is only $25\%$. Finally, when the person is not visible but close to the \textit{camera field-of-view} ($||\hvect_t - \Thetavect_t||_2<\tau$ and $\hvect_t\notin\mathcal{F}_t$), or when the person is unreachable ($\hvect_t\in[-\xi,\xi]\backslash[-1,1]$), this probability is $85\%$. Regarding the simulation of missing detections, we randomly ignore some faces when computing the face features. Concerning the sound modality, we randomly choose between the following cases: 1 person speaking, 2 persons speaking, and nobody speaking. We use a Markov model to enforce continuity in the speaking status of the persons, and we also simulate wrong audio observations.

From, the head position, we need to generate the position of all body joints. To do so, we propose to collect a set  $\mathcal{P}$
of poses from an external dataset (the AVDIAR dataset~\cite{gebru2017audio}). We use a multiple person pose estimator on this dataset and use the detected poses for our simulated environment. This task is not trivial since we need to simulate a realistic and consistent sequence of poses. Applying tracking to the AVDIAR videos could provide good pose sequences, but we would suffer from three major drawbacks. First, we would have a tracking error that could affect the quality of the generated sequences. Second, each sequence would have a different and constant size, whereas we would like to simulate sequences without size constraints. Finally, the number of sequences would be relatively limited. In order to tackle these three concerns, we first standardize the output coordinates obtained on AVDIAR. Considering the pose $p^n_t$ of the $n^{th}$ person, we sample a subset $\mathcal{P}^M_t \subset \mathcal{P}$ of $M$ poses. Then, we select the closest pose to the current pose: $p^n_{t+1}=\underset{p\in\Pi}{\argmin}~d(p,p^n_t)$ where
\begin{align}
  d\left(\begin{pmatrix}u_1\\v_1\\s_1\end{pmatrix},\begin{pmatrix}u_2\\v_2\\s_2\end{pmatrix}\right)=\frac{1}{\sum_{j=1}^J s_1^js_2^j} \sum_{j=1}^J (s_1^js_2^j)\sqrt{(u_1^j-u_2^j)^2+(v_1^j-v_2^j)^2}
\end{align}
This distance is designed to face poses with different number of detected joints. It can be interpreted as an $L_2$ distance weighted by the number of visible joints in common.
The intuition behind this sampling process is that when the size $M$ of $\mathcal{P}^M_t$ increases, the probability of obtaining a pose closer to $p^n_t$ increases. Consequently, the motion variability can be adjusted with the parameter $M$ in order to obtain a natural motion. With this method we can obtain diverse sequences of any size.

\begin{algorithm}[h]
  \KwData{$\mathcal{P}$: a set of poses, $\delta$: time-step \\
    $\sigma$: velocity variance, $M$: pose continuity parameter\vspace{0.3cm}}
 Randomly chose $N$ in $[1..3]$.\\
 \For{$n \in [1..N]$}{
 Initialize $(\hvect^n_0,\dot{\hvect}^n_0)\sim\mathcal{U}([-1,1])^2\times \mathcal{U}([-1,-0.5]\bigcup[0.5,1])^2$.\\
 Randomly chose $p^n_{0}$ in $\mathcal{P}$.\\
 }
 \For{$t \in [1..T-1]$}{
   \For{$n \in [1..N]$}{
     Randomly chose $motion\in\{Stay,Move\}$\\
       \eIf{$motion=Move$}{
         \eIf{$\hvect^n_t\notin[-\xi,\xi]^2$}{
           The person is leaving the scene.\\
           See section~\ref{subsec:simulated}.\\
         }{
           $\hvect^n_{t+1}\leftarrow\hvect^n_{t}+\delta(\dot{\hvect}^n_{t}+ \mathcal{N}((0,0),\sigma))$.\\
           $\dot{\hvect}^n_{t+1}\leftarrow\frac{1}{\delta}(\hvect^n_{t+1}-\hvect^n_{t})$\\
         }
       }{
         $\hvect^n_{t+1}\leftarrow\hvect^n_{t}$\\
         $\dot{\hvect}^n_{t+1}\sim\mathcal{U}([-1,-0.5]\bigcup[0.5,1])^2$\\
       }
  Draw $\mathcal{P}^M_t$, a random set of $M$ elements of $\mathcal{P}$\\
  $p^n_{t+1}\leftarrow\underset{p\in\mathcal{P}^M_t}{\argmin}~d(p,p^n_{t})$
   }
 }
 \caption{Generation of simulated moving poses for our simulated environment.}
\label{alg:synth}
\end{algorithm}

\section{Experiments}
\label{exper}
\begin{figure*}[t!]
  \centering
  \includegraphics[width=0.24\textwidth]{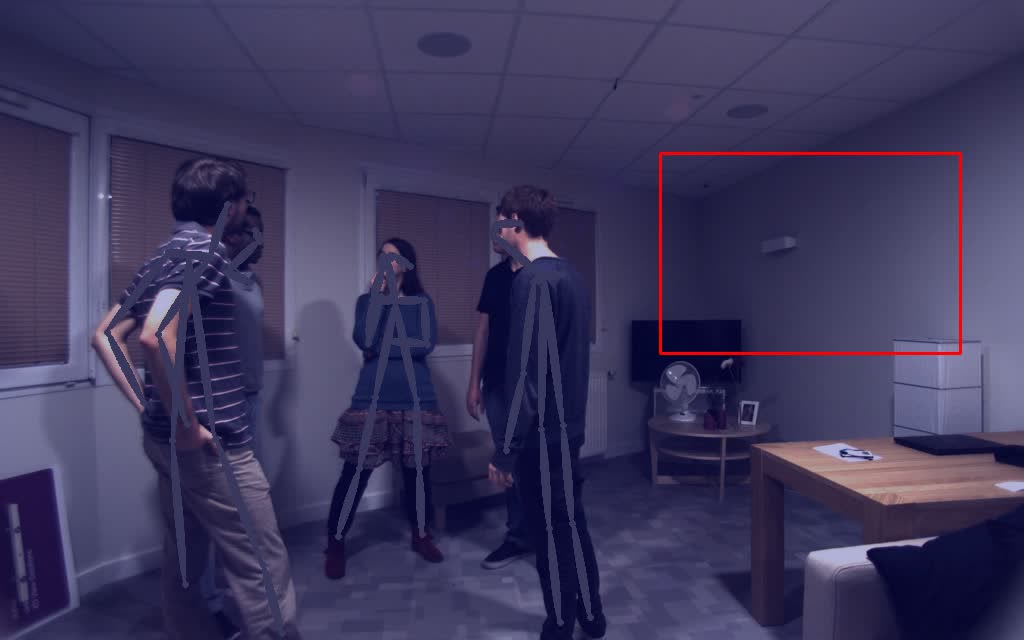}
  \includegraphics[width=0.24\textwidth]{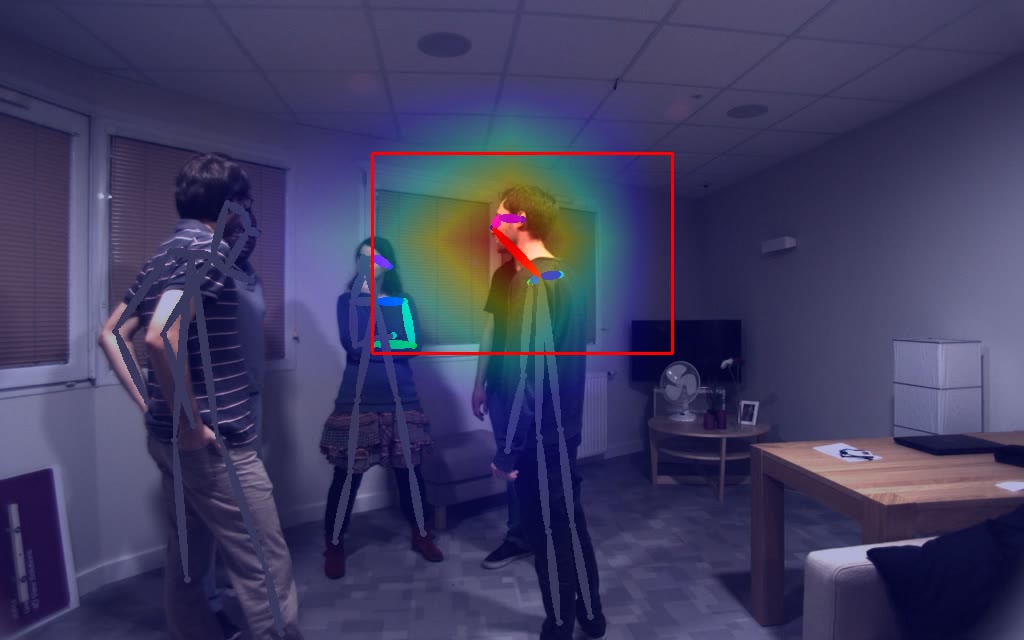}
  \includegraphics[width=0.24\textwidth]{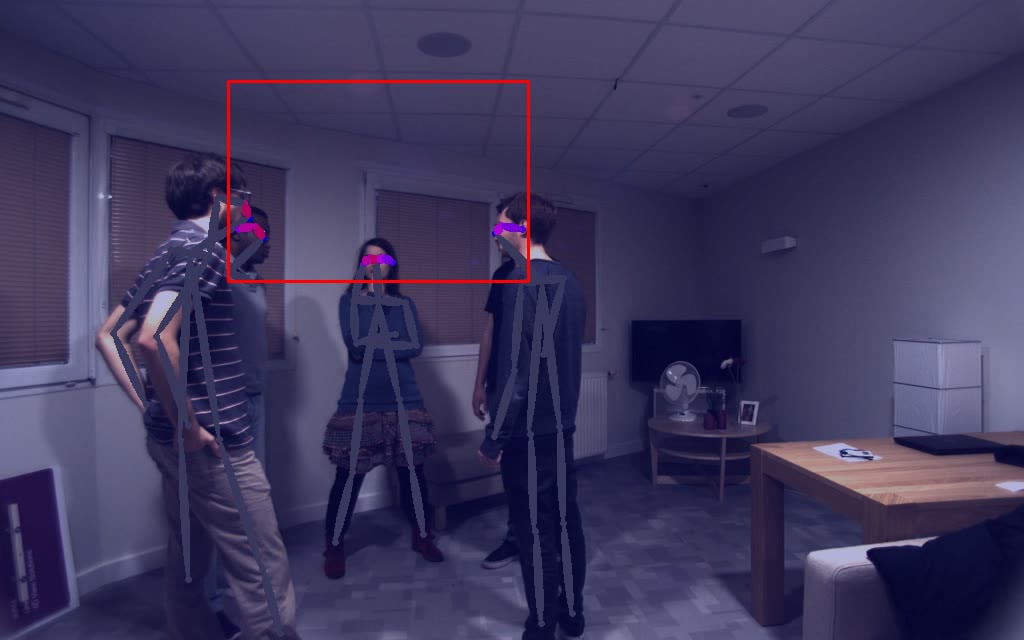}
  \includegraphics[width=0.24\textwidth]{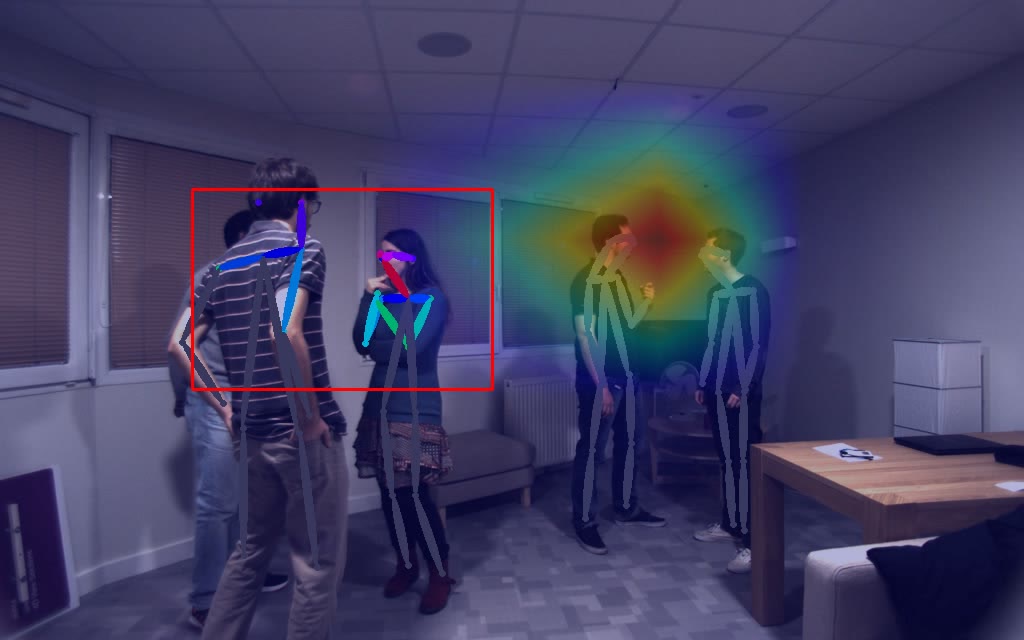}\\
  \caption{Example of a sequence from the AVDIAR dataset. The speech direction binary map is superimposed on the image, and the visible landmarks are displayed using a colored skeleton. The camera field of view (in red) is randomly initialized (far left), speech emitted by one of the persons is detected and hence the gaze is controlled (left). The agent is able to get all the persons in the field of view (right), and it gazes at a group of three persons while two other persons move apart (far right).}
  \label{fig:avdiarExpe}
\end{figure*}
\begin{figure*}[t!]
\centering
  \includegraphics[width=0.24\textwidth]{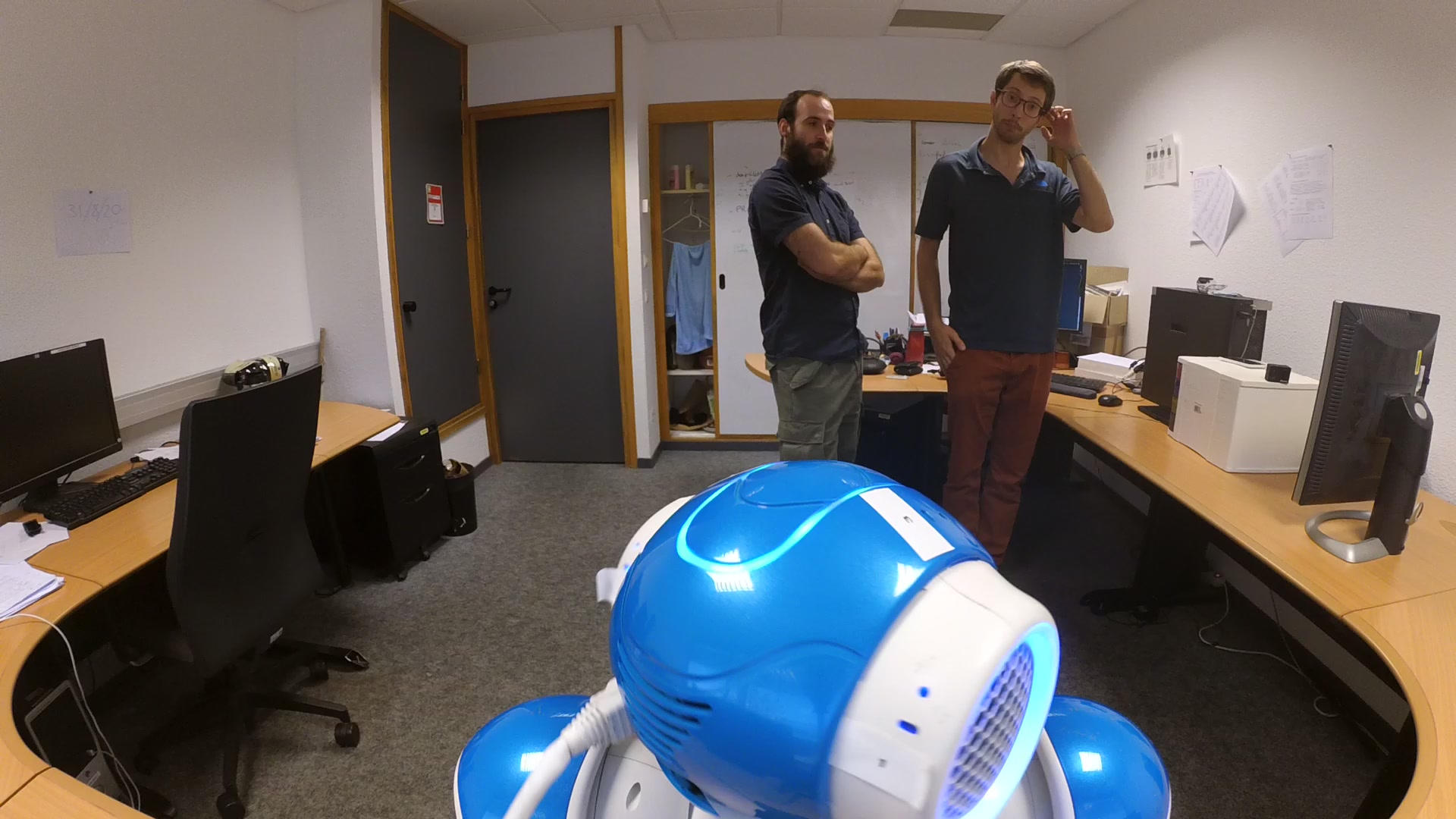}
  \includegraphics[width=0.24\textwidth]{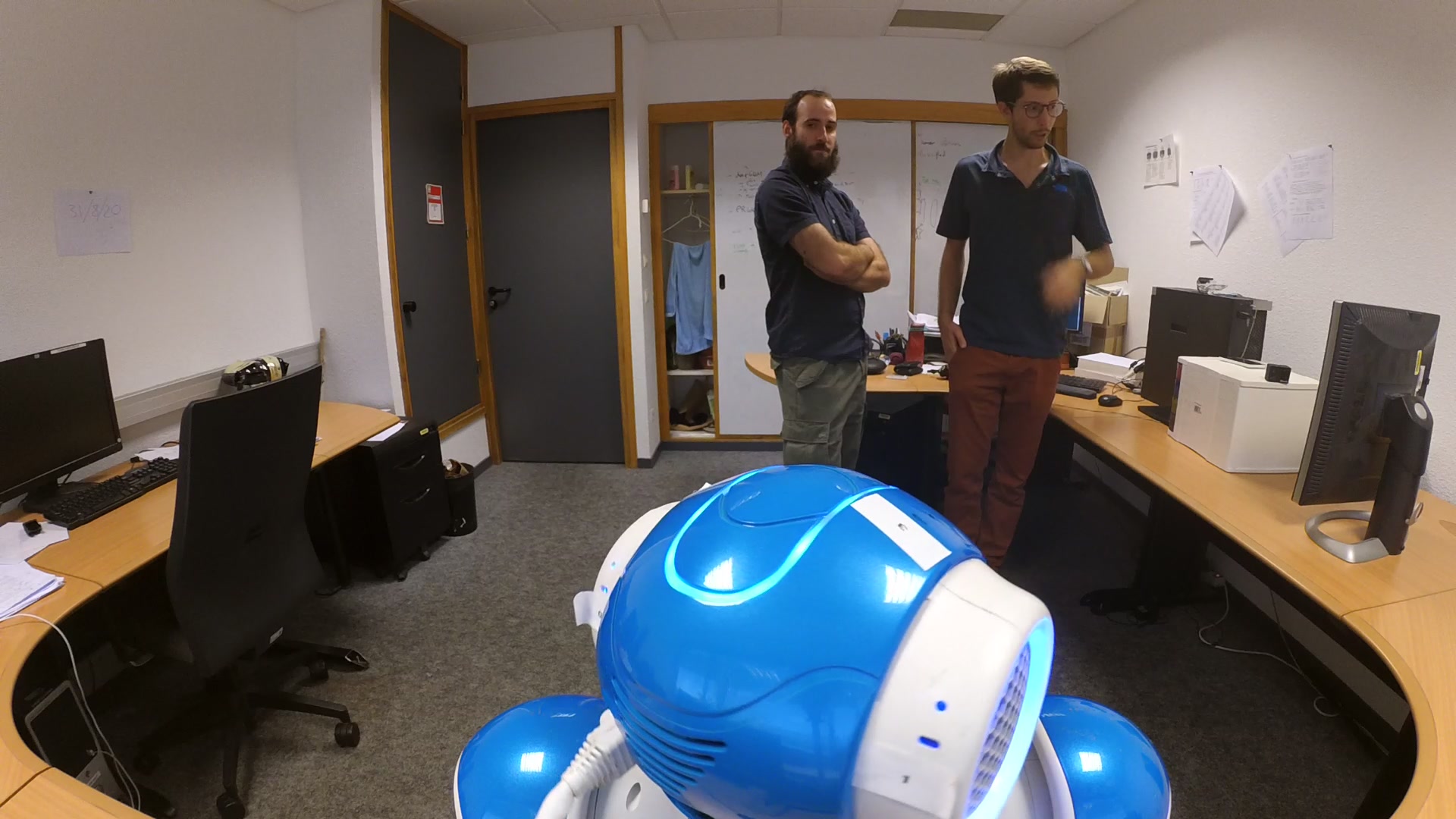}
  \includegraphics[width=0.24\textwidth]{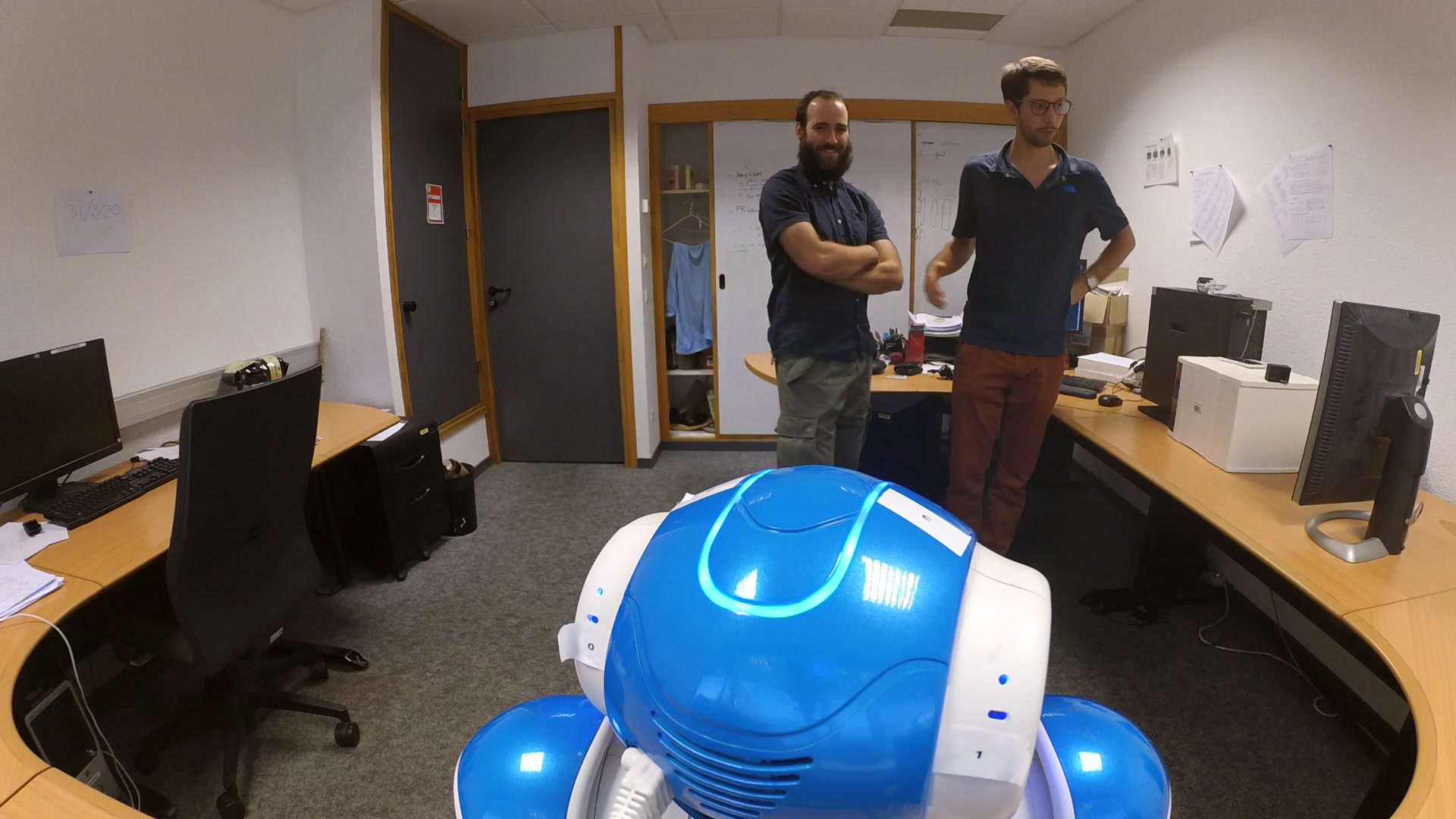}
  \includegraphics[width=0.24\textwidth]{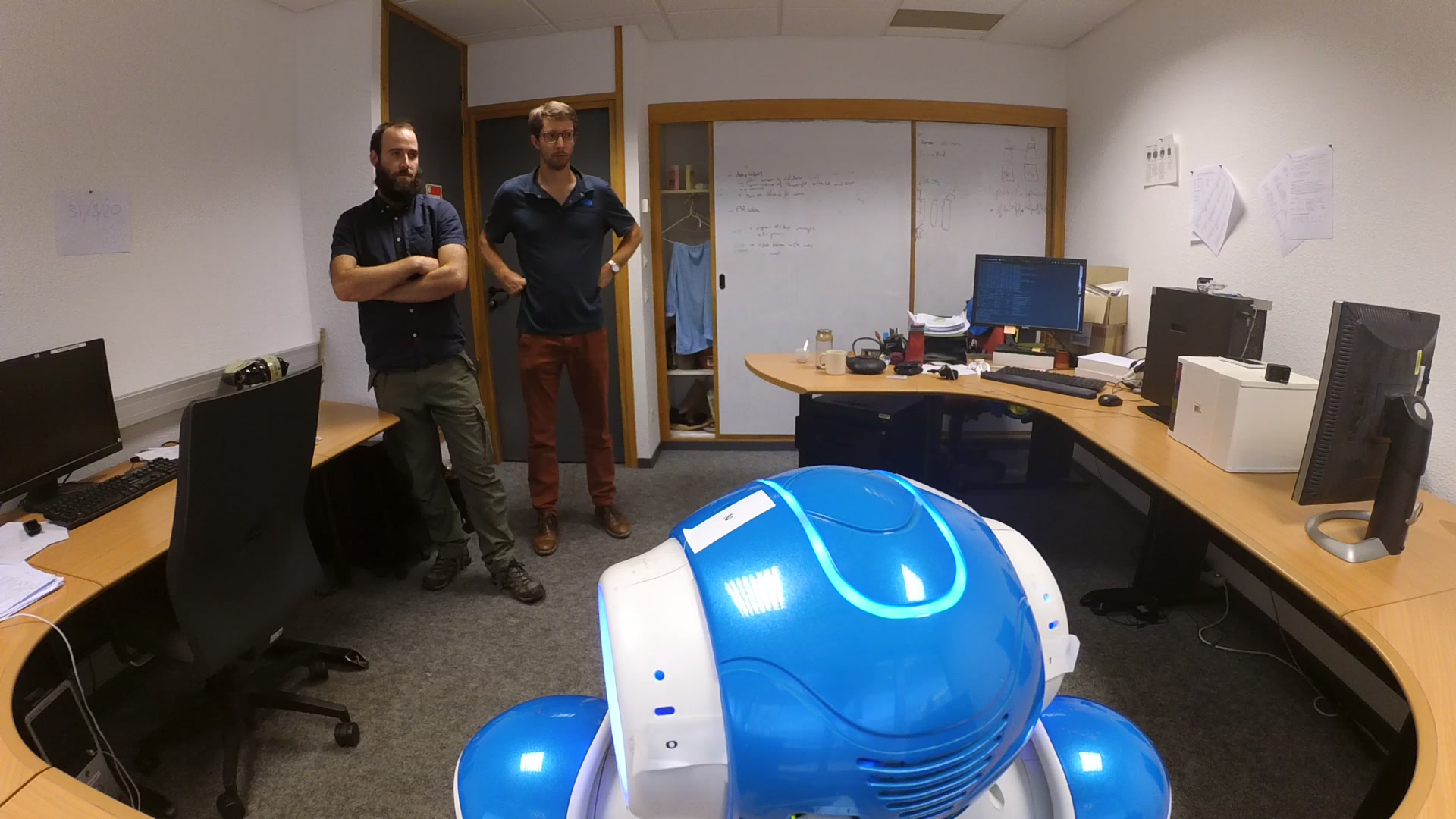}\\
  \includegraphics[width=0.24\textwidth]{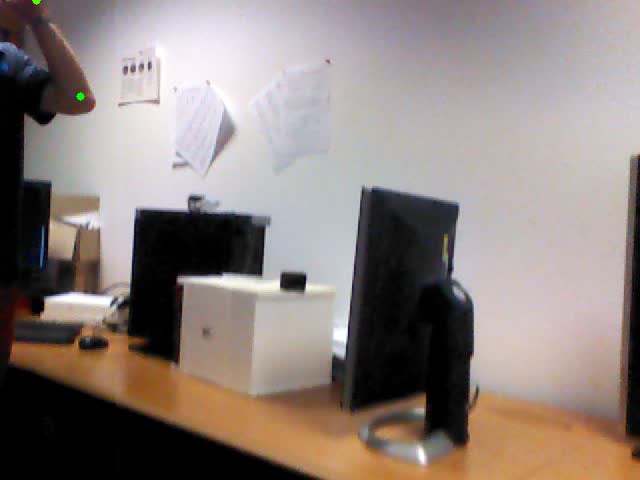}
  \includegraphics[width=0.24\textwidth]{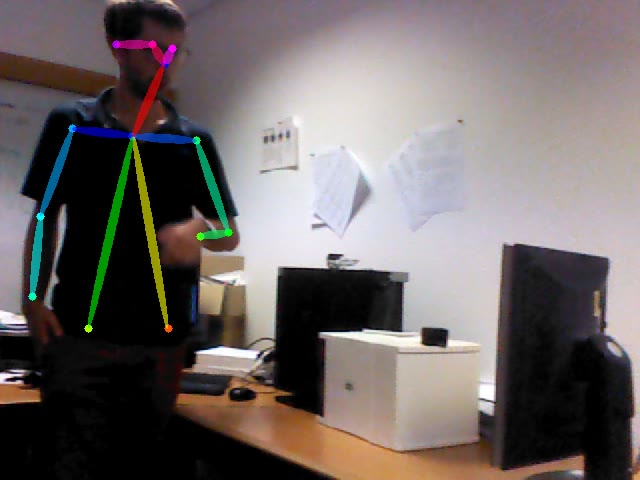}
  \includegraphics[width=0.24\textwidth]{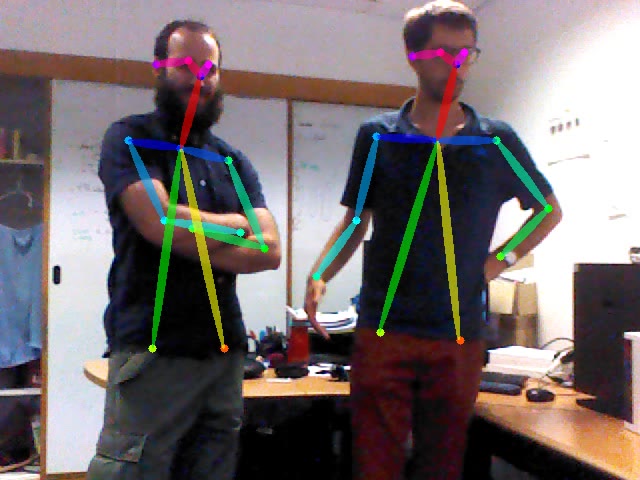}
  \includegraphics[width=0.24\textwidth]{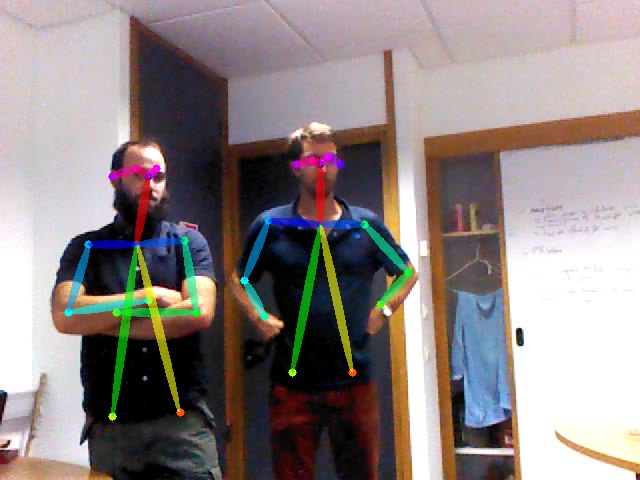}
  \caption{Example of a live sequence with two persons. First row shows an overview of the scene, including the participants and the robot. Second row shows the images gathered with the camera mounted onto the robot head. The robot head is first initialized in a position where no face is visible (first column), and the model uses the available landmarks (elbow and wrist) to find the person onto the right (second column). The robot detects the second person by looking around while keeping the first person in its field of view (third column), and gazes the two people walking together (fourth column).}
\label{fig:naoExpe}
\end{figure*}

\subsection{Evaluation with a Recorded Dataset}
\label{sub:AVDIAR}
The evaluation of HRI systems is not an easy task. In order to fairly compare  different models, we need to train and test the different models on the exact same data. In the context of RL and HRI, this is problematic because the data, i.e. what the robot actually sees and hears, depends on the action taken by the robot. Thus, we propose to first evaluate our model with the AVDIAR dataset~\cite{gebru2017audio}. This dataset was recorded with four microphones and one high-resolution camera ($1920\times1080$ pixels). These images, due to their wide field of view, are suitable to simulate the motor field of view of the robot. In practical terms, only a small box of the full image simulates the robot's camera field of view. \addnote[audiogridAVDIAR]{1}{Concerning the observations, we employ visual and audio grids of sizes $7\times 5$ in all our experiments with the AVDIAR dataset.}

We employ 16 videos for training. The amount of training data is doubled by flipping the video and audio maps. In order to save computation time, the original videos are down-sampled to $1024 \times 640$ pixels. The size of the camera field of view where faces can be detected is set to $300\times 200$ pixels using motion steps of 36 pixels each. These dimensions approximately correspond the coverage angle and motion of Nao. At the beginning of each episode, the position of the camera field of view is selected such that it contains no face. We noticed that this initialization procedure favors the exploration abilities of the agent. To avoid a bias due to the initialization procedure, we used the same seed for all our experiments and iterated three times over the 10 test videos (20 when counting the flipped sequences). An action is taken every 5 frames (0.2 seconds).

Figure~\ref{fig:avdiarExpe} shows a short sequence of the AVDIAR environment, displaying the whole field covered by the AVDIAR videos as well as the smaller field of view captured by the robot (the red rectangle in the figure).
\addnote[avdiarPB]{1}{However, it is important to highlight that transferring the model learned using AVDIAR to Nao is problematic and did not work in our preliminary experiments. First, faces are almost always located at the same position (around the image center). Second, all videos are recorded indoors using only two different rooms, and participants are not moving too much. Finally, the audio setting is unrealistic for a robotics scenario, e.g. absence of motor noise. Therefore, the main reason for using the AVDIAR dataset is to compare our method with other methods in a generic setting. }

%

\subsection{Live Experiments with Nao}
In order to carry out an online evaluation of our method, we performed experiments with a Nao robot. Nao has a $640\times480$ pixels cameras and four microphones. This robot is particularly well suited for HRI applications because of its design, hardware specifications and affordable cost. Nao's commercially available software can detect people, locate sounds, understand some spoken words, synthesize speech and engage itself in simple and goal-directed dialogs. Our gaze control system is implemented on top of the NAOLab middleware~\cite{badeig2015distributed} that synchronizes proprioceptive data (motor readings) and sensor information (image sequences and acoustic signals). The reason why we use a middleware is threefold. First, the implementation is platform-independent and, thus, easily portable. Platform-independence is crucial since we employ a transfer learning approach to transfer the model parameters, obtained with the proposed simulated environment, to the Nao software/hardware platform. Second, the use of external computational resources is transparent. This is also a crucial matter in our case, since visual processing is implemented on a GPU which is not available on-board of the robot. Third, the use of middleware makes prototyping much faster. For all these reasons, we employ the remote and modular layer-based middleware architecture named NAOLab. NAOLab consists of four layers: drivers, shared memory, synchronization engine and application programming interface (API). Each layer is divided into three modules devoted to vision, audio and proprioception, respectively. The last layer of NAOLab provides a general programming interface in C++ to handle the sensory data and to manage its actuators.  NAOLab provides, at each time step, an image
and the direction of the detected sound sources using \cite{li2017multiple,li2016reverberant}.


We now provide some implementation details specifically related  to the Nao implementation. The delay between two successive observations is $\sim$0.3 seconds.
 The rotating head has a motor field-of-view of $180^{\circ}$. The head motion parameters are chosen such that a single action corresponds to 0.15 radians ($\sim$9$\degree$) and 0.10 radians ($\sim$6$\degree$) for horizontal and vertical motions, respectively. \addnote[audiogridNao]{1}{Concerning the observations, we employ a visual grid of size $7\times 5$ and an audio grid of size $7\times 1$ in all our experiments with Nao. Indeed, Nao has a planar microphone array and hence sound sources can only be located along the azimuth (horizontal) direction. Therefore the corresponding audio binary map is one-dimensional.}

Figure~\ref{fig:naoExpe} shows an example of a two-person scenario using the \emph{LFNet} architecture. \addnote[commentNao]{1}{As shown in our recorded experiments
\footnote{A video showing offline and online experiments is available at \url{https://team.inria.fr/perception/research/deep-rl-for-gaze-control/}}
,  we were able to transfer the exploration and tracking abilities learned using the simulated environment. Our model behaves well independently of the number of participants.
    The robot is first able to explore the space in order to find people. If only one person is found, the robot follows the person. If the person is static, the robot keeps the previously detected person in the field but keeps exploring the space locally aiming at finding more people. When more people appear, the robot tries to find a position that maximizes the number of people. The main failure cases are related to quick movements of the participants.}

\begin{table*}[t!]
  \small
\caption{Comparison of the reward obtained with different architectures. The best results obtained are displayed in bold. }
\centering
\begin{tabular}{l|cc|cc|c|c}
  \hline
  &\multicolumn{4}{c|}{AVDIAR}&  \multicolumn{2}{c}{Simulated}\\
\hline
&\multicolumn{2}{c|}{Face}&\multicolumn{2}{c|}{Speaker}& \emph{Face} &\emph{Speaker}\\
Network&\emph{Training} & \emph{Test}&\emph{Training} & \emph{Test} &&\\
\hline
\emph{AudNet}&$1.50\pm 0.03$ & $1.47\pm 0.04$ &$1.92\pm 0.02$ & $1.82\pm 0.03$ & $0.21\pm 0.01$ & $0.33\pm 0.01$\\
\emph{VisNet}&$1.89\pm 0.03$ & $\bf 1.85\pm 0.02$ &$2.32\pm 0.04$ & $2.23\pm 0.03$ & $0.37\pm 0.04$ & $0.45\pm 0.06$\\
\emph{EFNet} &$1.90\pm 0.03$ & $1.81\pm 0.04$ &$2.40\pm 0.02$ & $2.22\pm 0.03$ & $0.41\pm 0.03$ & $\textbf{0.53}\pm \textbf{0.03}$\\
\emph{LFNet} &$\bf 1.96\pm 0.02$ & $1.83\pm 0.02$ &$\bf 2.43\pm 0.02$ & $\bf 2.29\pm 0.02$ & $\textbf{0.42}\pm \textbf{0.01}$ & $0.52\pm 0.03$\\
\hline
\end{tabular}
\label{tab:resultArch}
\end{table*}

\subsection{Implementation Details}
\label{implem}
By carefully selecting  the resolution used to perform person detection along the method of \cite{cao2017realtime}, we were able to obtain visual landmarks in less than 100~ms. Considering that NAOLab gathers images at 10~FPS, this landmark estimator can be considered as fast enough for our purpose. Moreover, \cite{cao2017realtime} follows a bottom-up approach, which allows us to speed-up landmark detection by skipping the costly association step.

The parameters of our model are based on a preliminary experimentation. We set $\Delta_T=4$ in all scenarios, such that each decision is based on the last 5 observations. The output size of  LSTM is set to 30 (since a larger size does not provide an improvement in performance), and the output size of the FCL is set to 5 (one per action). We use a discount factor ($\gamma$) of 0.90. Concerning the training phases, we employed the Adam optimizer~\cite{kingma2014adam} and a batch size of 128. In order to help the model to explore the policy space, we use an $\epsilon$-greedy algorithm: while training, a random action is chosen in $\epsilon\%$ of the cases; we decrease linearly the $\epsilon$ value from  $\epsilon=90\%$ to $\epsilon=10\%$ after 120000 iterations. 
The models were trained in approximately 45 minutes on both AVDIAR and the simulated environment. It is interesting to notice that we obtain this training time without using GPUs. A GPU is only needed  for person detection and estimation of visual landmarks (in our case, a Nvidia GTX 1070 GPU).

\begin{figure}[t!]
 \centering
 \subfloat[\emph{Face$\_$reward} on AVDIAR]{
   \includegraphics[width=0.40\columnwidth]{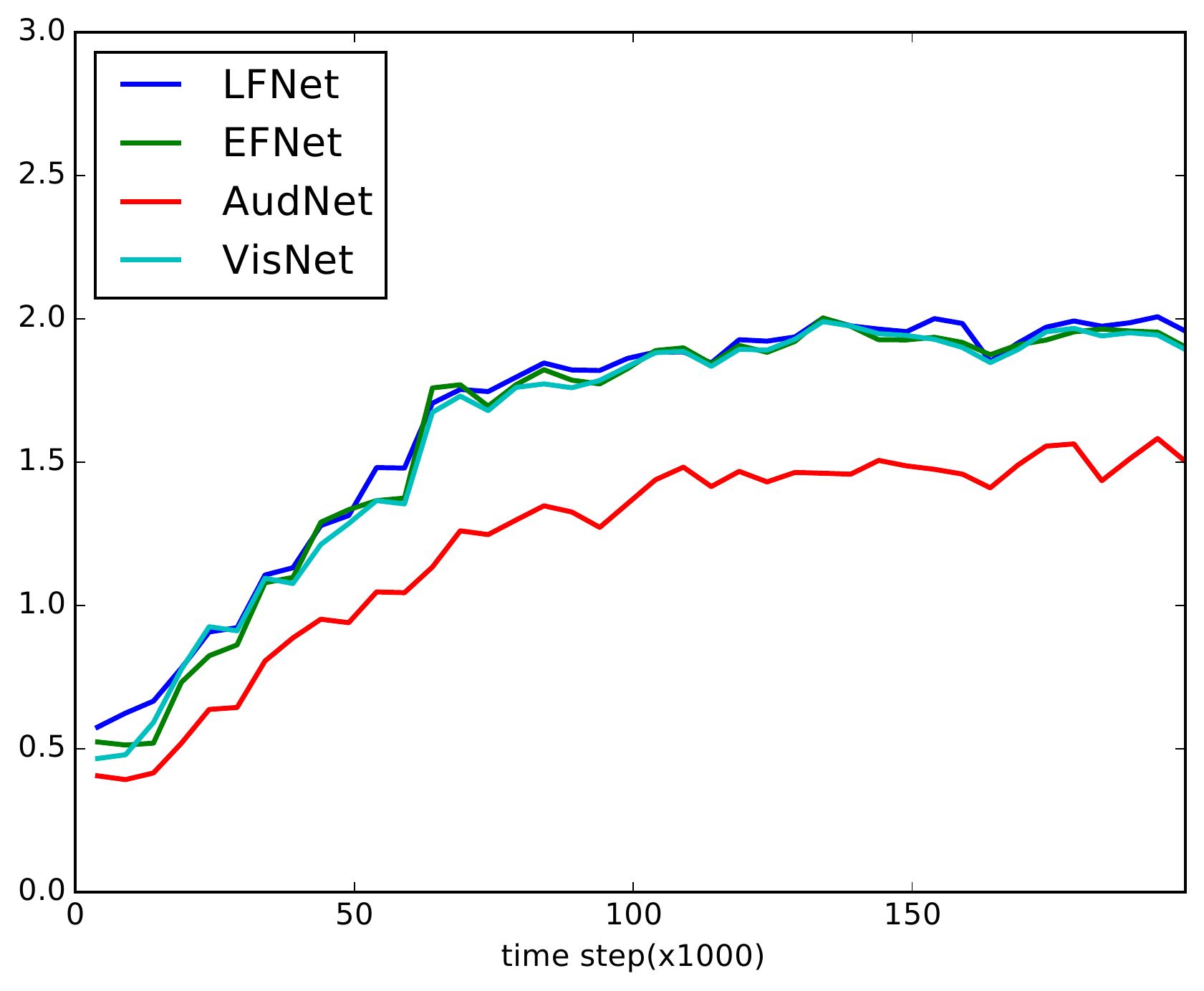}
   \label{fig:curveAVface}
 }
 \subfloat[\emph{Speaker$\_$reward} on AVDIAR]{
   \includegraphics[width=0.40\columnwidth]{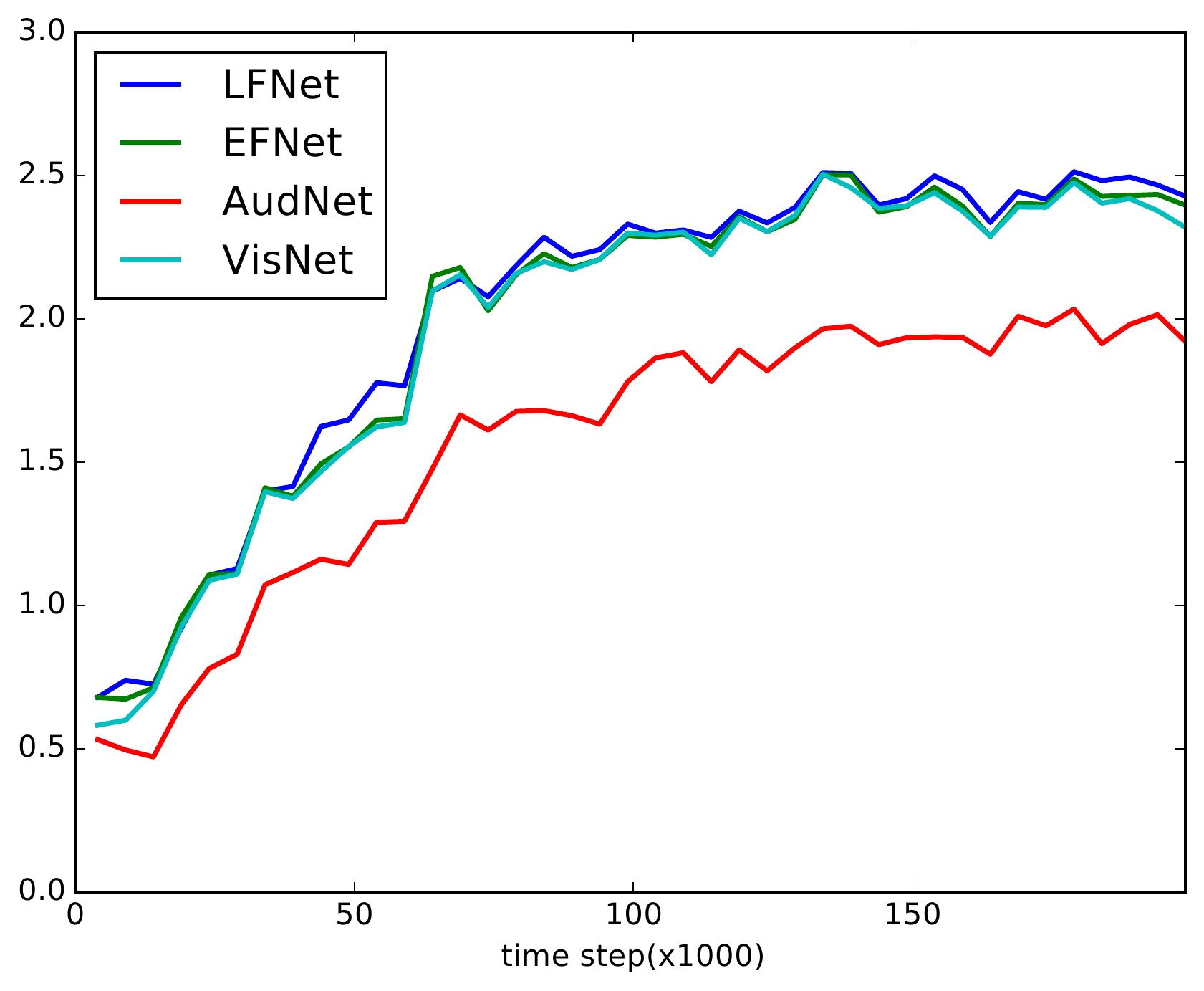}
   \label{fig:curveAVspe}
 }\\
 \subfloat[\emph{Face$\_$reward} on Simulated]{
   \includegraphics[width=0.40\columnwidth]{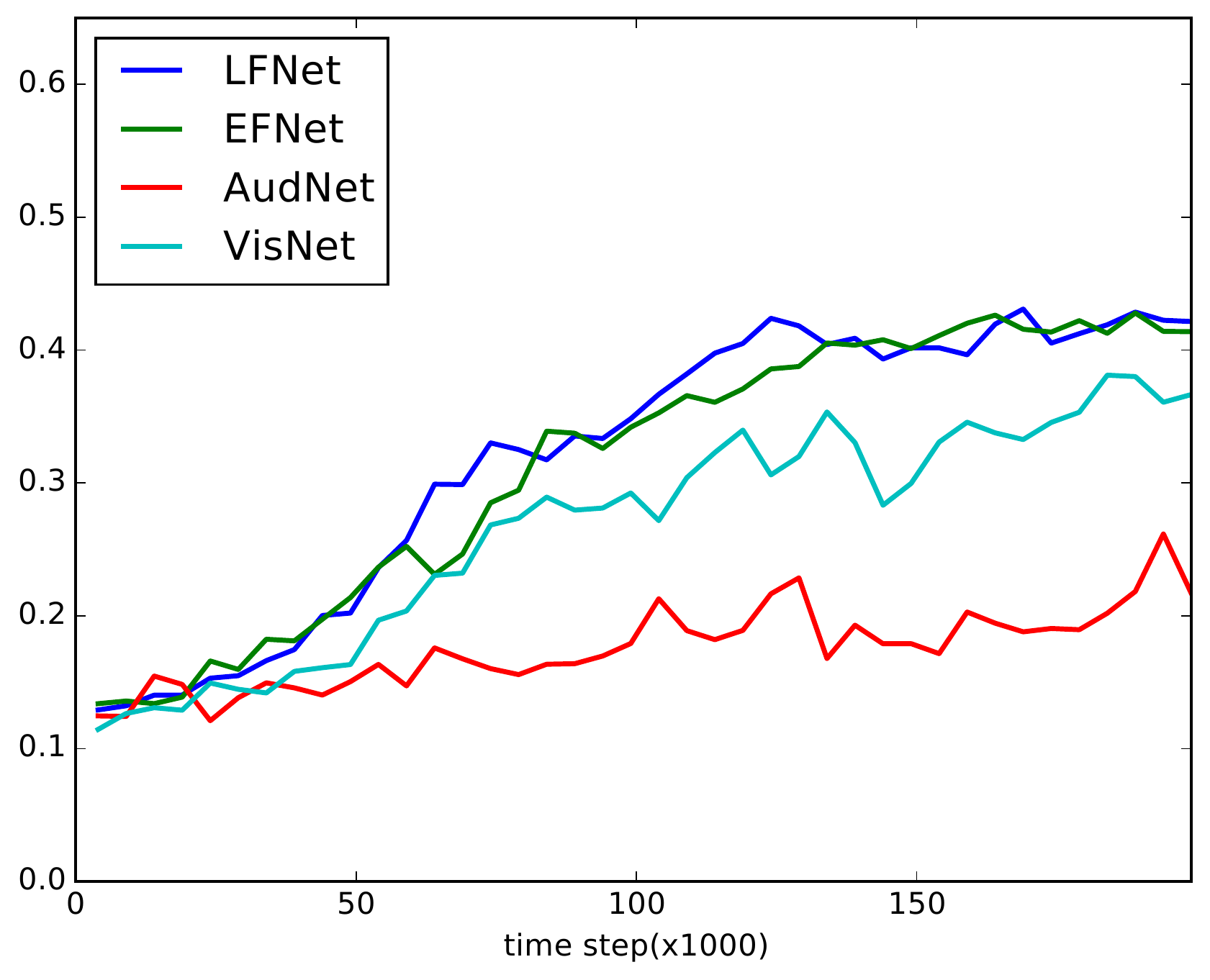}
   \label{fig:curveSynface}
 }
 \subfloat[\emph{Speaker$\_$reward} on Simulated]{
   \includegraphics[width=0.40\columnwidth]{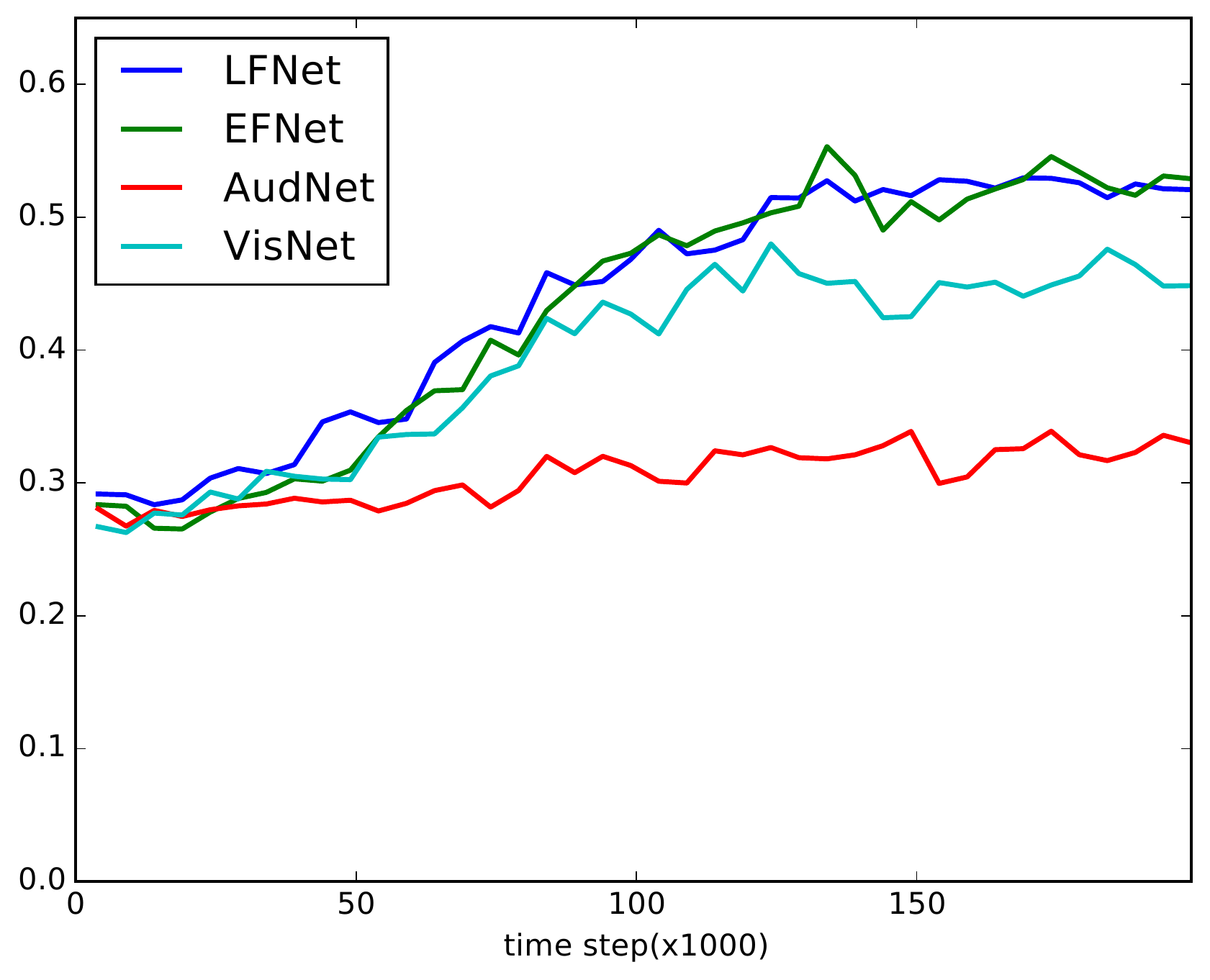}
   \label{fig:curveSynspe}
 }
 \caption{Evolution of the reward obtained while training with the two proposed rewards on the AVDIAR dataset and on the simulated environment. We average over a 5000 time-step window for a cleaner visualization.}
\label{fig:curve}
\end{figure}



In the simulated environment, the size of field in which the people can move is set to $\xi=1.4$. In the case of Nao, the audio observations are provided by the multiple speech-source localization method described in \cite{li2017multiple}.

In all our experiments, we run five times each model and display the mean of five runs to lower the impact of the stochastic training procedure.
On AVDIAR, the results on both training and test sets are reported in the tables. As described previously, the simulated environment is randomly generated in real time, so there is no need for a separated test set. Consequently, the mean reward over the last 10000 time steps is reported as test score.

\subsection{Architecture Comparison}

In Table \ref{tab:resultArch}, we compare the final reward obtained while training on the AVDIAR dataset and on our simulated environment with the two proposed rewards (\emph{Face$\_$reward} and \emph{Speaker$\_$reward}). Four different networks are tested: \emph{EFNet}, \emph{LFNet}, \emph{VisNet}, and  \emph{AudNet}. The y-axis of Figure \ref{fig:curve} shows the average reward per episode, with a clear growing trend as the training time passes (specially in the experiments with the AVDIAR dataset), meaning that the agent is learning (improving performance) from experience.
On the the simulated environment, the best results are indistinctly provided by the late and early fusion strategies (\emph{LFNet} and \emph{EFNet}), showing that our model is able to effectively exploit the complementarity of both modalities. On the AVDIAR, the late fusion performs slightly better than the early fusion model. Globally, we observe that the rewards we obtain on AVDIAR are higher than those obtained on the simulated environment. We suggest two possible reasons. First, the simulated environment has been specifically designed  to enforce exploration and tracking abilities. Consequently, it poses a more difficult problem  to solve. Second, the number of people in  AVDIAR  is higher (about 4 in average), thus finding a first person to track would be easier.  We notice that, on the AVDIAR dataset using the \emph{Face$\_$reward}, we obtain a mean reward greater than 1, meaning that, on average, our model can see more than one face per frame. 
 We also observe that \emph{AudNet} is the worst performing approach. However, it performs quite well on \emph{AVDIAR} compared to the simulated environment. This behavior can be explained by the fact that, on \emph{AVDIAR}, the speech source detector returns a 2D heatmap whereas only the  angle is used in the simulated environment. As conclusion, we select  \emph{LFNet}  to perform experiments on Nao.

\begin{figure}[h!]
  \centering
  \subfloat[AVDIAR]{
    \includegraphics[width=0.48\columnwidth]{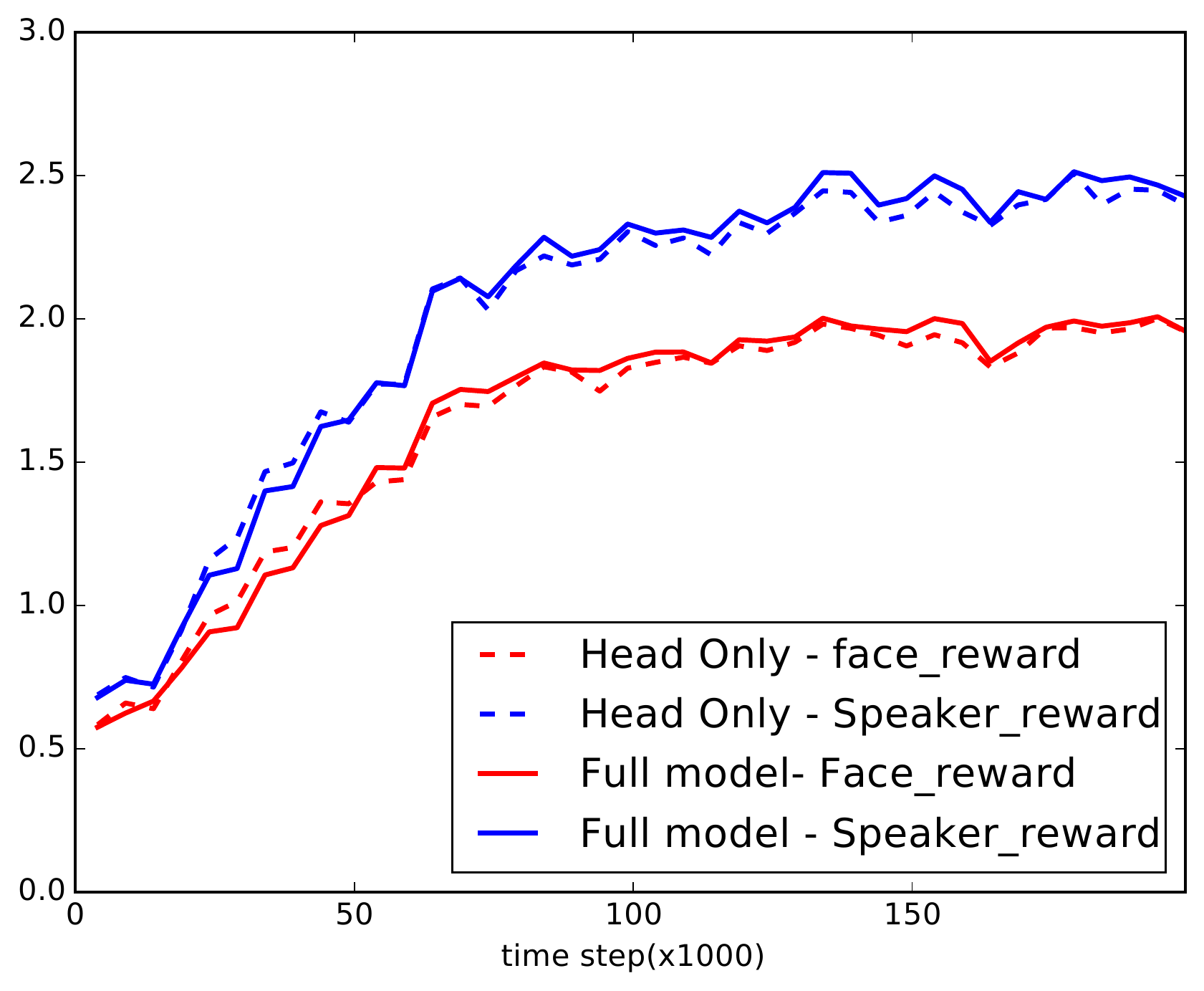}
    \label{fig:curveAVface}
  }
  \subfloat[Simulated]{
    \includegraphics[width=0.48\columnwidth]{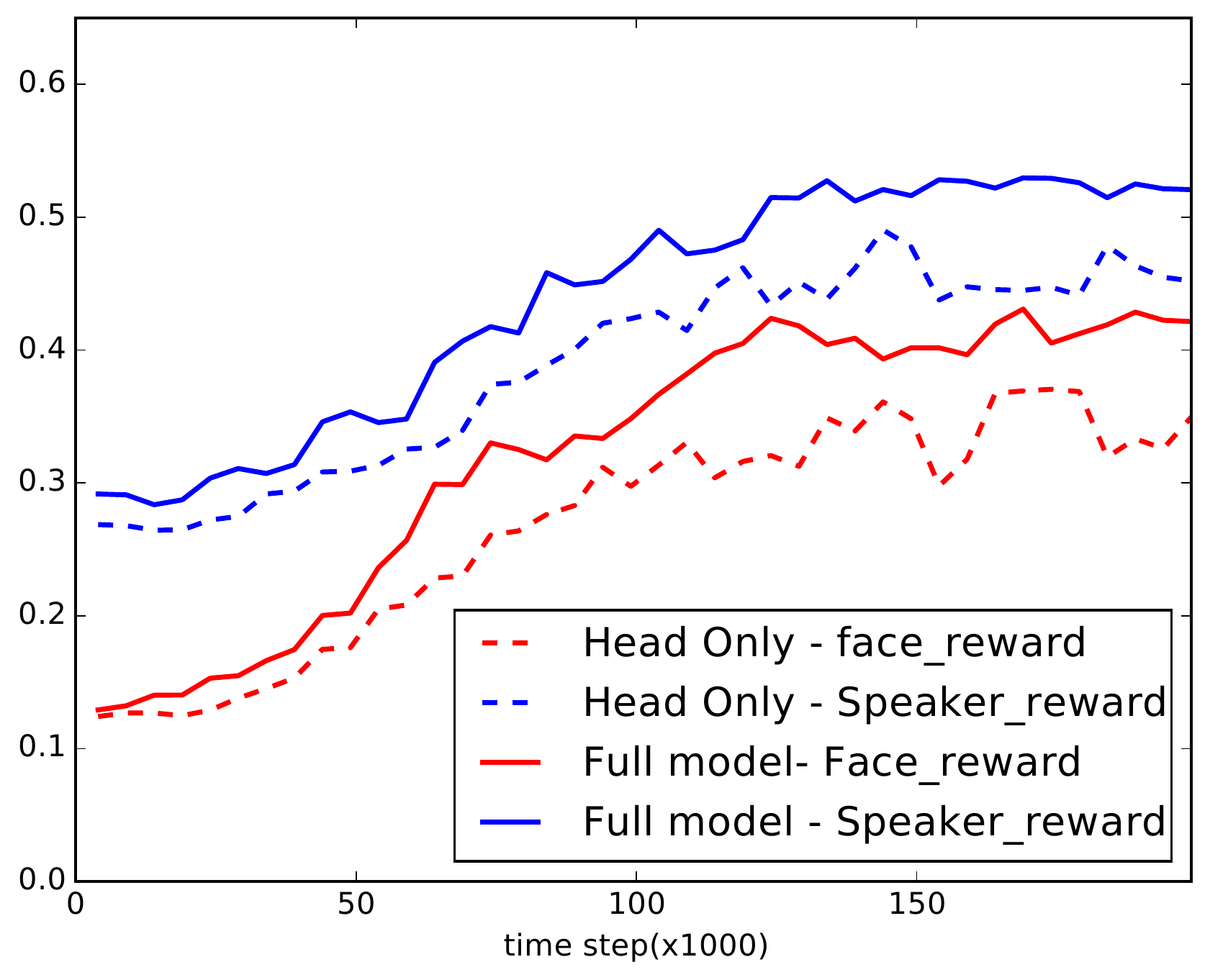}
    \label{fig:curveAVspe}
  }
  \caption{Evolution of the training reward obtained when using as visual observation the result of either the full-body pose estimation or the face location information.}
\label{fig:curveFullvsHead}
\end{figure}

Figure \ref{fig:curveFullvsHead} displays the reward obtained when  using only faces as visual observation (dashed lines) in contrast to using the full-body pose estimation (continuous lines). We observe that on the simulated data, the rewards are significantly higher when using the full-body pose estimator. This figure intends to respond empirically to the legitimate question of why a full-body pose estimator is used instead of a simple face detector. From a qualitative point of view, the answer can be found in the type of situations that can solve one and the other. Let's imagine that the robot looks at the legs of a user; in case of using only a face detector, there is no clue that could help the robot to move up its head in order to see a face; however, if a human full-body pose detector is used, the detection of legs implies that there is a torso over them, and a head over the torso.

\subsection{Parameter Study}

\begin{table*}[t!]
\caption{Comparison of the final reward obtained using different window lengths ($\Delta_T$). The mean and standard deviation over 5 runs are reported. The best average results obtained are displayed in bold. The training time is reported for each configuration.}
\centering
\begin{tabular}{c|ccc|cc}
  \hline
  &\multicolumn{3}{c|}{AVDIAR}&  \multicolumn{2}{c}{\emph{Simulated}}\\
$\Delta_T+1$&Training & Test & Time(s$\times10^3$) & Test & Time(s$\times10^3$)\\
  \hline
  1 & $1.92\pm 0.03$ & $1.82\pm 0.03$  & $3.05 \pm 0.22$& $0.26\pm 0.04$                   &$3.07 \pm 0.15$\\   
2 & $1.94\pm 0.02$ & $\bf 1.85\pm 0.02$ &$2.25 \pm 0.99$& $0.36\pm 0.04$                  &$3.09 \pm 0.17$\\  
3 & $1.93\pm 0.01$ & $1.84\pm 0.01$ &    $2.95 \pm 0.38$& $0.42\pm 0.02$                    &$2.98 \pm 0.27$\\  
5 & $1.94\pm 0.02$ & $1.84\pm 0.02$ &    $3.30 \pm 0.46$& $\textbf{0.43}\pm \textbf{0.01}$  &$3.40 \pm 0.14$\\  
10& $1.94\pm 0.02$ & $1.84\pm 0.02$ &    $2.05 \pm 0.22$& $0.40\pm 0.02$                    &$3.85 \pm 0.36$\\  
20& $\bf 1.96\pm 0.01$ & $1.82\pm 0.02$ &$3.00 \pm 0.00$& $0.42\pm 0.02$                 &$5.35 \pm 0.36$\\  
128&$1.94\pm 0.02$ & $1.82\pm 0.03$  &  $18.90 \pm 0.77$ & $0.41\pm 0.03$                      &$52.98 \pm 5.23$ \\
\hline
\end{tabular}
\label{tab:windowSize}
\end{table*}

 \addnote[paraComp]{1}{
  In this section, we describe the experiments devoted to evaluate the impact of some of the principal parameters involved. More precisely, the impact of three parameters is analyzed. First, we compare different values for the discount factor $\gamma$ that defines the importance of short-term rewards as opposed to long-term ones (see Section \ref{sec:model}). Second, we compare different window sizes. It corresponds to  the number of past observations that are used to make a decision (see Section \ref{QNetwork}). Finally, we compare different sizes for the LSTM network that is employed in all our proposed architectures (see Section \ref{QNetwork}). It corresponds to the dimension of the cell state and hidden state that are propagate by the LSTM.}

In Table \ref{tab:gamma}, different discount factors are compared. With AVDIAR, high discount factors are prone to overfit as the difference in performance between training and test is large. With the simulated environment, low discount values perform worse because the agent needs to perform several actions to detect a face, as the environment is rather complex. Consequently, a model that is able to take into account future benefits of each action performs better. Finally, in Table \ref{tab:LSTMSize}, we compare different LSTM sizes. We observe that increasing the size doesn't lead to better results, which is an interesting outcome since, from a practical point of view, smaller LSTMs faster the training.

Different window sizes  are compared in Table \ref{tab:windowSize}. We can conclude that the worst results are obtained when only the current observation is used (window size of 1). We also observe that, on AVDIAR, the model performs well even with short window lengths (2 and 3). In turn, with a more complex environment, as the proposed simulated environment, a longer window length tends to perform better. We interpret that using a larger window size helps the network to ignore the noisy observations and to remember the position of people that left the field of view. We report the training time for each window length. We observe that, using a smaller time window speeds up training since it avoids back-propagating the gradient deeply in the LSTM network.

\begin{table}[t!]
\caption{Comparison of the final reward obtained using different discounted factors ($\gamma$). The mean and standard deviation over 5 runs are reported. The best average results obtained are displayed in bold. }
\centering
\begin{tabular}{c|cc|c}
  \hline
  &\multicolumn{2}{c|}{AVDIAR}&  \emph{Simulated}\\
$\gamma$ &Training & Test &\\
  \hline
$ 25$ & $\bf 1.96\pm 0.02$ & $1.85\pm 0.02$ & $0.33\pm 0.09$ \\
$ 50$ & $\bf 1.96\pm 0.02$ & $\bf 1.86\pm 0.03$ & $0.35\pm 0.08$ \\
$ 75$ & $\bf 1.96\pm 0.02$ & $1.85\pm 0.02$ &  $\textbf{0.43}\pm \textbf{0.11}$ \\
$ 90$ & $1.94\pm 0.02$ & $1.83\pm 0.02$ & $0.42\pm 0.12$ \\
$ 99$ & $1.95\pm 0.01$ & $1.84\pm 0.02$ &  $0.42\pm 0.12$ \\
\hline
\end{tabular}
\label{tab:gamma}
\end{table}

\begin{table}[h]
\caption{Comparison of the final reward obtained using different LSTM sizes. The mean and standard deviation over 5 runs are reported. The best average results obtained are displayed in bold. }
\centering
\begin{tabular}{c|c|c|c}
  \hline
\emph{LSTM} size  &\multicolumn{2}{c}{AVDIAR}&  \emph{Simulated}\\
 &Training & Test &\\
\hline
$ 30$ & $\bf 1.96\pm 0.01$ & $1.85\pm 0.03$ & $0.42\pm 0.11$ \\
$ 60$ & $1.95\pm 0.02$ & $1.86\pm 0.02$  & $\textbf{0.43}\pm \textbf{0.12}$ \\
$ 120$ & $1.92\pm 0.04$ & $\bf 1.87\pm 0.02$ & $0.41\pm 0.10$ \\
\hline
\end{tabular}
\label{tab:LSTMSize}
\end{table}


\subsection{Comparison with the State of the Art}
\label{paraSOTA}
\addnote[noteSOTA1]{1}{
We perform a comparative evaluation with the state of the art. To the best of our knowledge, no existing work addresses the problem of finding an optimal gaze policy in the HRI context. In \cite{Bennewitz2005} a heuristic that uses an audio-visual input to detect, track and involve multiple interacting persons is proposed. Hence we compare our learned policy with their algorithm. On the simulated environment, as the speech source is only localized in the azimuthal plane (see section \ref{implem}), we randomly gaze  along the vertical axis in order to detect faces. In \cite{Ban17} two strategies are proposed to evaluate  visually controlled head movements. A first strategy consists of following a person and rotating the robot head in order to align the person's face with the image center. A second strategy consists in randomly jumping every 3 seconds between persons. Obviously, the second strategy was designed as a toy experiment and does not correspond to a natural behavior. Therefore, we compare our RL approach with their first strategy. Unfortunately, the case where nobody is in the field of view is not considered in \cite{Ban17}. To be able to compare their method in the more general scenario addressed here, we propose the following handcrafted policy in the case no face is detected in the visual field of view: (i)~\emph{Rand}: A random action is chosen; (ii)~\emph{Center}: Go towards the center of the \textit{acoustic field-of-view}; (iii)~\emph{Body}: If a limb is detected, the action $\uparrow$ is chosen in order to find the corresponding head, otherwise, \emph{Rand} is followed, and (iv)~\emph{Audio}: Go towards the position of the last detected speaker.
}
  
\addnote[noteSOTA2]{1}{
Importantly, in our model the motor speed is limited, since the robot can only select unitary actions. When implementing other methods, one could argue that this speed limitation is inherent to our approach and that other methods may not suffer from it. However, it is not realistic to consider that the head can move between two opposite locations of the auditory field in two consecutive frames with an infinite speed. Therefore, we report two scores, first using the same speed value than the one used in our model (referred to as \emph{equal}), and second by making the unrealistic assumption that the motor speed is infinite (referred to as \emph{infinite}). This second evaluation protocol is therefore biased  towards handcrafted methods. The results are reported in Table \ref{tab:compSOA}.
}

\begin{table*}[t!]
  \small
\caption{Comparison of the rewards obtained with different handcrafted policies. The performances of competitor methods are reported considering the two speed assumptions (\emph{equal}/\emph{infinite}) described in the text.}
\centering
\begin{tabular}{l|cc|cc}
  \hline
  &\multicolumn{2}{c|}{AVDIAR}&  \multicolumn{2}{c}{Simulated}\\
\hline
&\emph{Face$\_$reward}&\emph{Speaker$\_$reward}& \emph{Face$\_$reward} &\emph{Speaker$\_$reward}\\
\hline
Ban et al.\cite{Ban17}+\emph{Rand}  & $1.19/1.21 $ & $1.45/1.59$ &$0.25/0.26  $  &$0.40/0.37$\\
Ban et al.\cite{Ban17}+\emph{Center}& $1.62/1.68 $ & $1.95/2.01$ &$0.14/0.11  $  &$0.28/0.29$\\
Ban et al.\cite{Ban17}+\emph{Body}   & $1.23/1.20$& $1.40/1.52 $ &  $0.27/0.26 $&$0.39/0.37 $\\  
Ban et al.\cite{Ban17}+\emph{Audio}    & $1.54/1.63$& $1.84/2.06 $ &  $0.32/0.39 $&$0.43/0.48 $\\
Bennewitz et al.\cite{Bennewitz2005} &$1.56/1.55 $ &$2.07/2.05$& $0.30/$ $\bf0.42 $& $0.35/0.50$ \\
\emph{LFNet} &$\bf1.83\pm 0.02$ & $\bf 2.29\pm 0.02$ & $\textbf{0.42}\pm \textbf{0.01}$ & $\bf0.52\pm 0.03$\\
\hline
\end{tabular}
\label{tab:compSOA}
\end{table*}

\addnote[noteSOTA3]{1}{
First, we notice that none of the handcrafted methods can compete with ours when considering the same motor speed. On both environments, \emph{LFNet} largely outperforms all handcrafted models. This clearly justifies policy learning and the use of RL for gaze control. Concerning \cite{Ban17}, \emph{Center} obtains the best result among the \cite{Ban17}'s variances on AVDIAR and the worst on Simulated according to the \emph{Face\_reward} metric. This can be explained by the fact that, as mentioned in Section \ref{sub:AVDIAR}, most persons are located around the image center and, therefore, this dummy strategy works better than more sophisticated ones. A similar behavior can be observed with the \emph{Speaker\_reward} metric. We observe that in both environments using audio information, when no face is detected, improves the performance with respect to \emph{Rand}. The second best performance on AVDIAR is obtained by  \cite{Bennewitz2005} with \emph{Speaker\_reward}. On the simulated environment,  \cite{Bennewitz2005} equals the score obtained by our proposal when making the unrealistic assumption of infinite motor speed. In that case,  \cite{Bennewitz2005} is marginally inferior to our proposal according to the \emph{Speaker\_reward}. When considering equal speed limit, our RL approach significantly outperforms  the handcrafted policy of \cite{Bennewitz2005} ($26\%$ and $48\%$ higher according to \emph{Face\_reward} and \emph{Speaker\_reward}, respectively).
All these results highlight the crucial importance of audio-visual fusion in the framework of RL and in the context of gaze control. 
}

\section{Conclusions}
In this paper, we presented a neural network-based reinforcement learning approach to solve the gaze robot control problem. In particular, our agent is able to autonomously learn how to find people in the environment by maximizing the number of people present in its field of view while favoring people that speak. A simulated environment is used for pre-training prior to transfer learning to a real environment. Neither external sensors nor human intervention are necessary to compute the reward. Several architectures and rewards are compared on three different environments: two offline (real and simulated datasets) and real experiments using a robot. Our results suggest that combining audio and visual information leads to the best performance, as well as that pre-training on simulated data can even make unnecessary to train on real data. By thoroughly experimenting on a publicly available dataset and with a robot, we provide empirical evidence that our RL approach outperforms handcrafted strategies. 
\section*{Acknowledgments}
EU funding through the ERC Advanced Grant VHIA \#340113 is greatly acknowledged.

\bibliographystyle{abbrv}

\end{document}